\documentclass[sn-mathphys-num]{sn-jnl}% Math and Physical Sciences Numbered Reference Style 
\usepackage{float}
\usepackage[table,xcdraw]{xcolor} % Include xcolor package for row colors
\usepackage{adjustbox}
\usepackage{graphicx}%
\usepackage{multirow}%
\usepackage{amsmath,amssymb,amsfonts}%
\usepackage{amsthm}%
\usepackage{mathrsfs}%
\usepackage[title]{appendix}%
\usepackage{xcolor}%
\usepackage{textcomp}%
\usepackage{manyfoot}%
\usepackage{booktabs}%
\usepackage{algorithm}%
\usepackage{algorithmicx}%
\usepackage{algpseudocode}%
\usepackage{listings}%
\usepackage{adjustbox}
\def\BibTeX{{\rm B\kern-.05em{\sc i\kern-.025em b}\kern-.08em
    T\kern-.1667em\lower.7ex\hbox{E}\kern-.125emX}}
\theoremstyle{thmstyleone}%
\theoremstyle{thmstyletwo}%

\theoremstyle{thmstylethree}%

\begin{document}

\title[Article Title]{Toward a universal concept of artificial personality: implementing robotic personality in a Kinova arm}


\author*[1,2]{\fnm{Alice} \sur{Nardelli}}\email{alice.nardelli@edu.unige.it}
\author[3]{\fnm{Lorenzo} \sur{Landolfi}}
\author[2]{\fnm{Dario} \sur{Pasquali}}
\author[1]{\fnm{Antonio} \sur{Sgorbissa}}
\author[2]{\fnm{Francesco} \sur{Rea}}
\author[1]{\fnm{Carmine} \sur{Recchiuto}}

\affil[1]{\orgdiv{Department of Informatics, Bioengineering, Robotics, and Systems Engineering}, \orgname{ University of Genoa}, \orgaddress{\street{Via all'Opera Pia 13}, \city{Genoa}, \postcode{16150}, \state{Italy}, \country{Italy}}}

\affil[2]{\orgname{COgNiTive Architecture for Collaborative Technologies (CONTACT) Unit, Italian Institute of Technology (IIT)}, \orgaddress{\street{Via Enrico Melen}, \city{Genoa}, \postcode{16150}, \state{Italy}, \country{Italy}}}

\affil[3]{\orgname{Unit for Visually Impaired People (U-VIP), Italian Institute of Technology (IIT)}, \orgaddress{\street{Via Enrico Melen}, \city{Genoa}, \postcode{16150}, \state{Italy}, \country{Italy}}}

\abstract{The fundamental role of personality in shaping interactions is increasingly being exploited in robotics. A carefully designed robotic personality has been shown to improve several key aspects of Human-Robot Interaction (HRI). However, the fragmentation and rigidity of existing approaches reveal even greater challenges when applied to non-humanoid robots. On one hand, the state of the art is very dispersed; on the other hand, Industry 4.0 is moving towards a future where humans and industrial robots are going to coexist. In this context, the proper design of a robotic personality can lead to more successful interactions. This research takes a first step in that direction by integrating a comprehensive cognitive architecture built upon the definition of robotic personality—validated on humanoid robots—into a robotic Kinova Jaco2 arm. The robot personality is defined through the cognitive architecture as a vector in the three-dimensional space encompassing Conscientiousness, Extroversion, and Agreeableness, affecting how actions are executed, the action selection process, and the internal reaction to environmental stimuli.
Our main objective is to determine whether users perceive distinct personalities in the robot, regardless of its shape, and to understand the role language plays in shaping these perceptions. To achieve this, we conducted a user study comprising 144 sessions of a collaborative game between a Kinova Jaco2 arm and participants, where the robot’s behavior was influenced by its assigned personality. Furthermore, we compared two conditions: in the first, the robot communicated solely through gestures and action choices, while in the second, it also utilized verbal interaction.

}


\maketitle

\section{Introduction}\label{sec:intro}

\begin{figure}[h]
\centering
\includegraphics[width=0.4\textwidth]{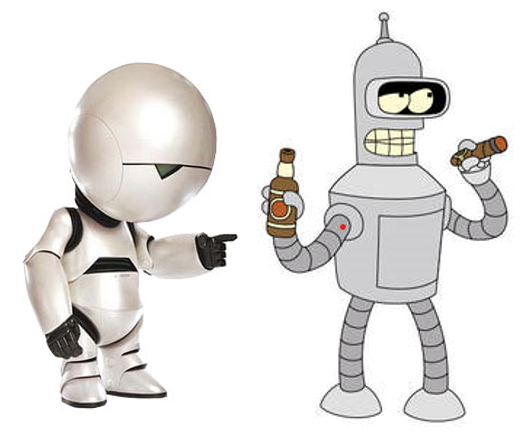}
\caption{Marvin the paranoid android, and Bender the aggressive robot.}\label{fig:marvin}
\end{figure}

Personality reveals itself through patterns of thoughts, behaviors, and feelings \cite{diener2019personality}, and consequently, it shapes individual and group dynamics. Large-scale population research demonstrates that several key aspects of relationships—such as marriages, job performance, and friendship networks—are influenced by personality traits \cite{mount1998five}, \cite{russell1991personality}, \cite{correa2010interacts}. As our society moves towards a future where humans and robots coexist, it is important to consider how personality can influence Human-Robot Interaction (HRI).

Popular imagination has already begun to envision relationships between robots, artificial agents, and humans shaped by robotic personalities. For instance, consider the unconventional, pointless, and irritating Genuine People Personalities from the famous novel \textit{The Hitchhiker's Guide to the Galaxy} \cite{adams1995hitch}, where the tricky relationship between the paranoid android Marvin—who is always depressed—and the extremely cheerful and sunny doors serves as an example. Similarly, the animated science fiction series \textit{Futurama} features Bender, a hedonistic and aggressive robot, as a protagonist  (Figure \ref{fig:marvin} from \footnote{\url{https://www.deviantart.com/pxz5pm/art/Marvin-CS-S-151433443}} and \footnote{\url{https://www.deviantart.com/superawesomevectors/art/Futurama-Bender-Vector-Character-646313289}}).

A tailored design of robotic personality has proven to be an effective strategy for improving various aspects of interaction, such as likeability, enjoyment, knowledge acquisition, engagement, trust, and empathy \cite{peeters2006personality}, \cite{robert2018personality}, \cite{esterwood2021systematic}, \cite{robert2020review}, \cite{rossi2020role}. Additionally, robotic personality has been shown to be a viable approach for persuading users in decision-making \cite{paradeda2020persuasion}, \cite{dangelo23}, mitigating the novelty effect \cite{gockley2005designing}, and improving the appeal of interactions by leveraging the similarity-complementarity theory \cite{Lim2022538}, \cite{Aly2016193}, \cite{martins2020i2e}.

The growing necessity and importance of HRI in the Industry 4.0 sector is reflected in emerging research trends that aim to integrate cognitive capabilities into industrial robots \cite{landolfi2023working}, \cite{rojas2019variational} to improve mutual understanding and psychological ergonomics \cite{gualtieri2020opportunities}. A carefully considered design of an industrial robot's behavioral parameters, potentially influenced by personality, can enhance collaboration, psychological safety, and trust \cite{landolfi2023working}, \cite{story2022speed}, \cite{brule2014robot}, \cite{porubvcinova2020determinants}. However, systematic reviews on robotic personality research \cite{mou2020systematic} reveal that the vast majority of robotic personality implementations have been in humanoid robots, with a notable absence of research on implementing personality in robotic arms.

In this work, we propose the integration of a task- and platform-independent cognitive architecture to implement robotic personality \cite{nardelli2024personality}, \cite{nardelli2024ei} in a Jaco2 Kinova robotic arm. The cognitive architecture draws inspiration from human psychology to define a proper taxonomy addressing the limitation of using a single trait, typically extroversion, to describe a robotic personality. We propose the CEA (Conscientiousness, Extroversion, and Agreeableness) taxonomy of robotic personality \cite{nardelli2023software}, based on three traits of the Big Five factor model \cite{goldberg1981language}. Additionally, we address the limitations of rigid robotic personality implementations that are typically tied to specific tasks and platforms. To this aim, we implement personality as an abstract concept by exploiting the generalizability of BERT (Bidirectional Encoder Representations from Transformers) attention-based architecture \cite{devlin2018bert}, fine-tuning it to predict a comprehensive set of parameters that apply across various tasks and platforms influenced by personality. Furthermore, our proposed architecture bridges the gap between cognitive agents by integrating memory capabilities, prospective thinking, and emotional intelligence. This integration recognizes that personality affects not only the execution of actions but also the decision-making process and the hedonic experience associated with those actions.

Given these premises, our first objective is, therefore, to determine whether a combination of the CEA traits, managed by the aforementioned cognitive architecture, can be perceived in a robotic arm, which will validate the task and platform independence of the cognitive architecture, not only on humanoid robots \cite{nardelli2023software}, \cite{nardelli2024personality} and digital humans \cite{nardelli2024ei} but also on industrial robots. Secondly, we aim to explore the impact of language on the perception of robotic personality in a robotic arm. 

These two objectives are addressed through a user study in which participants engage in a collaborative game with the Kinova arm. We compare two conditions: in the first, the robot's personality is expressed only through movement parameters and action decision-making; in the second, the robot is also able to express itself through language.

It is important to highlight that this work represents a pioneering implementation of artificial personality in a robotic arm and may be a first step toward understanding how robotic personality can enhance HRI in Industry 4.0, where humans and robots are increasingly required to coexist and collaborate.

\section{State of the Art}
\label{sec:ssa}

Personality is crucial for modeling, understanding, and predicting human behaviors and emotions \cite{peeters2006personality}. Incorporating social capabilities such as emotions and personalities into robots engages the human emotional sphere, leading to greater acceptance \cite{bartneck2006use} and increased empathy \cite{riek2009anthropomorphism}.

Systematic reviews of robotic personality research \cite{mou2020systematic} indicate that most implementations are designed for humanoid robots. However, a growing body of research also focuses on non-humanoid robots. Several studies have explored implementing robotic personalities to understand their impact on user engagement and preferences. For example, Mower et al. \cite{mower2007investigating} implemented three distinct personalities—neutral, positive, and negative—in a Pioneer 2DX mobile robot to examine how personality traits, expressed through vocal cues, language, and camera movements, affected user performance during a wire puzzle game. They found that user errors decreased when interacting with the robot’s negative personality.

In a post-stroke rehabilitation context, Tapus and colleagues \cite{tapus2008user} designed introvert and extrovert personalities in a mobile robot by modulating proxemics, speed, and vocal content. In the first experiment, they tested a fixed personality, while the second and third experiments involved an adaptive personality. Their findings support the similarity-attraction hypothesis, as extroverted users preferred interacting with an extroverted robot personality, finding it more engaging. Additionally, they demonstrated that adaptive robot behaviors tailored to user personality and performance were more effective in the rehabilitation activity.

Groom et al. \cite{groom2009my} investigated perceived traits, such as friendliness, integrity, and malice, in a Lego car robot and a humanoid Lego-assembled robot. The study’s conditions varied based on whether the robot was self-assembled or other-assembled and whether it was humanoid or car-like, within a game activity. Participants showed greater attachment to the car robot, which they rated as friendlier and having more integrity. In contrast, the humanoid robot was perceived as more independent and even malicious. Users felt a greater sense of ownership over the robot they had assembled themselves.

Kim et al. \cite{kim2009entertainment} examined the influence of motion factors—direction, speed, volume, and repetition—on perceived personality dimensions of friendliness and dominance using the ROLLY robot. They found that speed and direction positively impacted perceived dominance, while repetition, speed, and direction positively affected perceived friendliness. However, volume negatively affected friendliness.

Hendriks et al. \cite{hendriks2011robot} focused on user preferences for vacuum cleaner personalities based on the Big Five traits, implementing preferred traits through modulations in motion (trajectory, velocity, acceleration, regularity, force), light (on-off, speed, regularity), and sound (volume, pitch, timbre). Their results showed that people anthropomorphize robot vacuum cleaners and attribute personality characteristics to them. Users preferred a calm, polite, and cooperative vacuum cleaner that operated efficiently and systematically, reflecting preferences for reliability and routine.

This fragmented state-of-the-art highlights that research on robotic personalities in non-humanoid robots has been largely neglected in the last decade. Furthermore, the implementation of robotic personality in robotic arms is even rarer, as most research has concentrated on mobile platforms.

Previous studies \cite{nardelli2023software}, \cite{nardelli2024personality}, \cite{nardelli2024ei} identify several issues with robotic personality research that become even more pronounced in the context of non-humanoid robots. Firstly, the literature lacks an effective taxonomy of personality \cite{Luo2022}. As noted in review activities \cite{mou2020systematic}, robotic personality research, regardless of the robot's shape, tends to focus primarily on extroversion. Defining personality effectively in terms of traits is challenging when considering non-humanoid robots. 
Recently \cite{nardelli2023software}, we have proposed a model for defining and implementing robotic personalities based on three dimensions of the Big Five (Consciousness, Extroversion, Agreableness, CEA), which has been extensively tested with humanoid robots, but, in line of principle, could be used on any robotic platform. In this work, we aim to implement this taxonomy on a robotic arm further to demonstrate its generalizability across different types of robots.

Secondly, the literature indicates that personality is often expressed multimodally \cite{andriella2022know},\cite{Aly2016193}, influencing various channels such as language generation, vocal cues, facial expressions, and body language. However, a comprehensive set of behavioral parameters adaptable to different platforms has not yet been identified. In this work, we aim to highlight the universality of the parameters identified in \cite{nardelli2023software} by applying them to industrial robots. Among the works reviewed \cite{mower2007investigating}, \cite{tapus2008user}, \cite{groom2009my}, \cite{kim2009entertainment}, \cite{hendriks2011robot} it is possible to distinguish between those that convey personality in non-humanoid robots through a combination of movement and speech parameters and those that use only movement parameters. This research seeks to determine whether gestures alone are sufficient to convey personality in a robotic arm and to assess the impact of language parameters on the perception of personality.

Last, personality also affects the cognitive system \cite{vernon2014artificial}, including memory encoding, prospective thinking, hedonic experiences, internal responses to others' perceptions, action-selection processes, and emotional intelligence. This connection is only partially addressed in simulations within the affective computing field \cite{kotseruba202040}, \cite{cabrera2022adaptive}, \cite{bourgais2020ben}, \cite{martins2020i2e}. The literature on robotic personality has largely overlooked these aspects, especially in the context of industrial robotics. However, it is important to emphasize that both artificial personality and cognitive capabilities are deeply intertwined. Implementing these capabilities can enhance interaction and mutual understanding between humans and robots, thereby improving psychological ergonomics. These aspects are fundamental in Industry 4.0, which is guided by the paradigm of coexistence. Although this research focuses solely on the perception of personality, it represents an initial step toward smoother interactions by implementing the task- and platform-independent cognitive architecture proposed in \cite{nardelli2023software}, \cite{nardelli2024personality}, \cite{nardelli2024ei}, which aims to bridge the gap between psychological and cognitive agents also in the industrial robotic field.

\section{Material and Methods}
\label{sec:mmm}

\subsection{Cognitive Architecture}
\label{sec:mm}

\begin{figure}[h]
\centering
\includegraphics[width=0.9\textwidth]{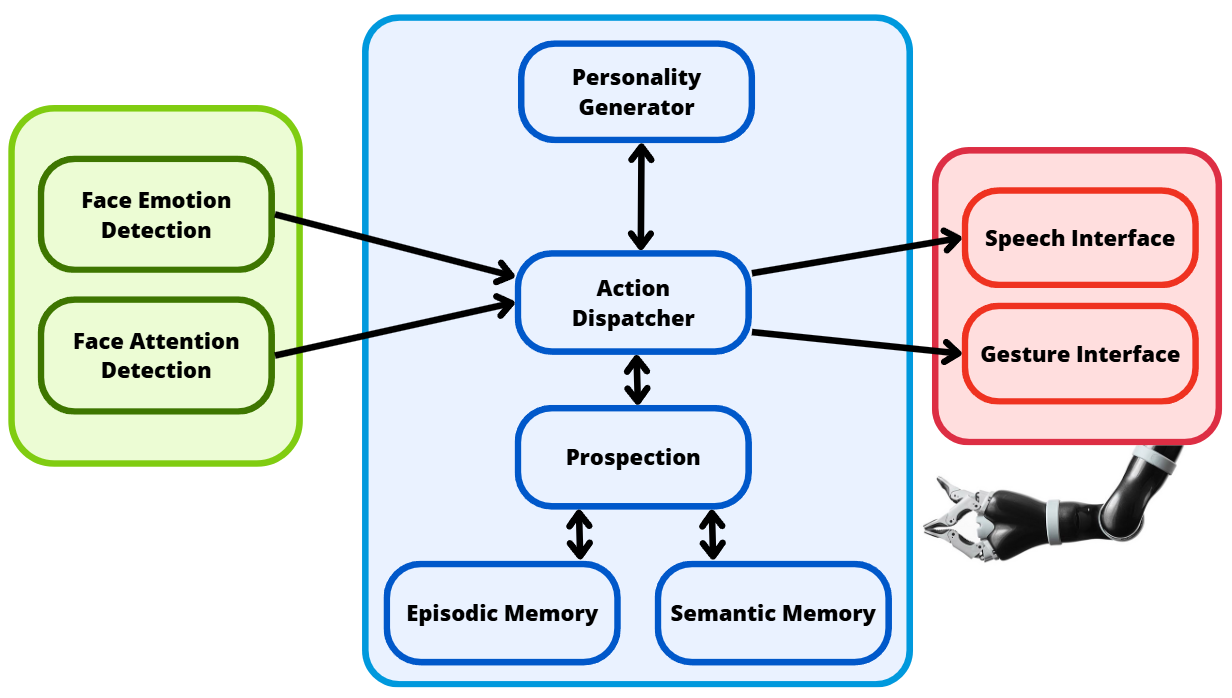}
\caption{Cognitive Software Architecture for the Kinova Jaco2 robotic arm. The diagram identifies the cognitive components in blue, the perception components in green, and the action components in red.}\label{fig:sofar}
\end{figure}

In this research, we adapted the cognitive architecture proposed in \cite{nardelli2023software}, \cite{nardelli2024personality}, \cite{nardelli2024ei} to a Kinova Jaco2 arm \cite{kinova1} (Figure \ref{fig:sofar}), to assess the capability of the cognitive architecture to generate distinguishable personalities even in a non-humanoid robot.

Below, the main components of the cognitive architecture and their adaptation to the manipulator robot are described. With reference to Figure 2, the green perception blocks estimate user emotion and attention from facial expressions through Morphcast software \cite{morphcast}. We categorize emotions according to Ekman's six basic emotions: Happy, Sad, Surprise, Disgust, Fear, and Anger \cite{ekman1992there}, plus the Neutral state, that we have considered to identify also the situation in which no specific emotions are shown by the user. To mitigate outliers, we consider the most probable emotion within a fixed sliding window of 3 seconds. Given that facial expressions can be deceptive, implementing a multimodal emotion recognition system, as done in \cite{nardelli2024ei}, can significantly improve the accuracy of emotion estimation. It is worth saying, however, that our focus was on implementing a robotic personality in a robotic arm rather than achieving precise emotion detection. 

%Dealing with the Board Perception module, we utilized the OpenCV library \cite{opencv1} to detect the chessboard and LEGO blocks.

The thinking and emotional components of the overall architecture, dedicated to implementing robotic personality, are depicted in blue. To address the complexity of human personality and simplify it into a practically applicable model for HRI contexts, we define a taxonomy for artificial personality tailored for robotics \cite{nardelli2023software}. The CEA taxonomy represents robotic personality as a vector in a three-dimensional space defined by Conscientiousness, Extroversion, and Agreeableness (Section \ref{sec:taxonomy}). This approach allows for the concurrent management of multiple traits, overcoming the common limitation of using a single trait to describe robotic personalities.

The Personality Generator is implemented by fine-tuning a BERT (Bidirectional Encoder Representations from Transformers) attention-based architecture \cite{devlin2018bert}, and it is designed to simulate personality-influenced behaviors (Section \ref{sec:psycho}). Its implementation and finetuning procedure is presented in detail in our previous research \cite{nardelli2023software}.

Establishing connections between the Personality Generator and the cognitive system (see Sections \ref{sec:psycho} and \ref{sec:prospection}) enables us to simulate how personality affects the memory system, internal simulations, emotional intelligence, hedonic feelings, and, consequently, the action selection process.

The framework captures the ability to describe past events through natural language, characteristic of Semantic Memory \cite{vernon2014artificial}, using an ontology. This component stores a personality-independent description of the world using a set of propositions and predicates, which can be retrieved and recalled when necessary.

Episodic Memory encodes the link between past episodes of a specific action and the associated reward, in terms of comfortability, obtained by executing it. This allows the agent to choose future actions that maximize the personality-dependent hedonic experience. The cognitive capability involved in this personality-dependent, goal-directed behavior is known as Prospection \cite{vernon2014artificial}. This enables the agent to simulate future actions and their associated comfort by learning from past episodes, their meanings, and their outcomes. This is achieved by linking Prospection directly with memory functions. We implement Prospection using a Fast-Forward (FF) planner \cite{hoffmann2001ff}. Its numerical planning capabilities allow us to simulate the personality-dependent comfort function that needs to be maximized in the action-selection process. Additionally, the planner is integrated into the system to enable iterative planning in response to action failures, changes in the environment due to new perceptions, or actions' outcomes that differ from expectations (Section \ref{sec:prospection}).

The Action Dispatcher orchestrates the overall system by managing the flow of information between different components. It processes input to update the system on the current task state, along with the user's emotional and attentional states. The direct connection between the Action Dispatcher and Prospection allows the system to determine the action to execute basing on the effect that the task-dependent actions and the perceived emotional and attentional states have on comfortability. For example, a social robot with an agreeable or extroverted personality involved in a collaborative task might ask the user to perform the task, as this action involves social interaction. In contrast, if the robot has an introverted or disagreeable personality, it might choose to carry out the task directly. 
Additionally, once a new emotional or attentional state is perceived, the comfort level within Prospection is updated, which can trigger new actions. For example, an empathic collaborative robot that notices a user feeling sad might feel uncomfortable. This could trigger a comfort-driven action, such as encouraging the user before asking them to continue the task. This enables the agent to be proactive within the environment and execute actions that depend only on comfortability.
%Additionally, the Action Dispatcher queries the Game Player to accurately understand the new move (e.g., placing the block in area 2 of the chessboard).

Broadly, the Action Dispatcher updates Prospection and queries it when a new action needs to be executed. At that time, the Action Dispatcher retrieves from the Personality Generator the behavioral parameters required for that specific action and triggers the action block.

%Since the Personality Generator predicts inactive parameters for actions not needed, the system automatically detects the action block to trigger. 

%Regarding the red action blocks, the Gesture Interface is implemented through the Kinova arm's API, allowing the robot to perform tasks as well as communicative actions (Section \ref{sec:gesture}).

Regarding the red action blocks, the Gesture Interface executes the robot’s movements to perform tasks dependent actions as well as communicative actions. All these actions are properly coded with the parameters generated by the Personality Generator (Section \ref{sec:gesture}). Conversely, the Speech Interface generates the appropriate sentence following the approach presented in our previous research \cite{nardelli2024ei}. 

%Conversely, the Speech Interface is used to produce a sound wave to notify the human when it is their turn to place a block on the chessboard if the robot communicates solely through gestures. When the system can also communicate verbally, GPT-4o model \cite{chatgpt} is queried to generate the appropriate sentence. The text generation process is detailed in our previous research \cite{nardelli2024ei}. The Microsoft Cognitive System \cite{azure} is then used to synthesize the voice stream.

The overall architecture utilizes ROS middleware to facilitate communication between components, a choice made to ensure the system's adaptability to various industrial robotic platforms.

\subsection{CEA taxonomy}
\label{sec:taxonomy}
The literature on robotic personalities primarily focuses on extroversion (Section \ref{sec:ssa}) and a model of personalities applicable to all robots, regardless of size, shape, and purpose, does not currently exist \cite{Luo2022}.
The work performed in \cite{luo2022identifying}, which specifically investigates artificial personalities for humanoid robots based on the Big Five model \cite{goldberg1981language}, reveals that humans can identify only three personality dimensions: Extroversion, Conscientiousness, and Agreeableness. This conclusion is supported by the findings in \cite{Luo2022}, which indicate that humans can distinguish between 4 to 8 personalities, likely corresponding to various combinations of the two extremes of these three dimensions. Conversely, the research in \cite{volkel2022user}, which examines how the extroversion trait is perceived during human-robot interactions, suggests that extroversion is also interpreted through the lenses of Conscientiousness and Agreeableness. This highlights the limitations of using a single trait to define a robot's personality. All these aspects, together with experimental data deriving from tests with humanoid robots \cite{nardelli2023software}, \cite{nardelli2023software}, \cite{nardelli2024personality}, suggest the usage of our three-dimensional models with non humanoid robots as well. Indeed, the resulting CEA taxonomy describes the agent's personality as a vector in the three-dimensional space of Conscientiousness, Extroversion, and Agreeableness traits drawn from the Big Five factor model \cite{goldberg1981language}, which are the ones more easily perceived by users during interactions with robots \cite{luo2022identifying}. Consequently, it is possible to define infinite personalities managing multiple traits at a time, described as follow:

$$
Personality= W_{c}C +  W_{e}E +  W_{a}A \eqno{(1)}
$$ 

where \(C\), \(E\), and \(A\) correspond to the versors of the three axes, and \(W_{c}\), \(W_{e}\), and \(W_{a}\) are the corresponding coordinates of the specific personality vector representing the degree to which a specific trait is expressed. The three coordinates \(W_{c}\), \(W_{e}\), and \(W_{a}\) vary in the range \([-1, +1]\), with \(0\) associated to a neutral personality.

\subsection{Personality Generator}
\label{sec:psycho}
Personality affects how we execute actions. Robotic literature primarily focuses on this aspect for the implementation of robotic personalities; however, each study tends to focus on a limited set of parameters specific to the robot or task, resulting in rigid implementations that do not allow for universal conclusions about personality-driven behavioral parameters. In humans, personality is not dependent on the specific person or the task they are engaged in; rather, it encompasses universal traits that influence a wide range of behavioral tendencies, which may not always be expressed depending on the actions performed.

To address this issue, our previous research \cite{nardelli2023software} proposed a set of behavioral parameters that are not dependent on the specific task or platform but are influenced by the Conscientiousness, Agreeableness, and Extroversion traits. The resulting parameters affect aspects such as Vocal Cues, Language Style, Gaze, Head Movements, Gestures, and Navigation.

In the following, we focus on the parameters related to Vocal Cues, Language, and Gestures, as these aspects are the only ones applicable to a manipulator, i.e., the robotic platform considered for the experimental evaluation. For a more comprehensive description of the mapping between personality and parameters, the reader can refer to \cite{nardelli2023software}.

We followed the methodology proposed by \cite{polzehl2015personality} to map between Vocal Cues and CEA traits. %In conditions where the robot only moves, the volume parameter is used to produce sound waves. In contrast, when the robot is in speaking mode, the volume parameter is used to reproduce the audio stream.

Personality is also expressed through verbal markers \cite{de2010only}, \cite{saucier1999hierarchical}, \cite{john1988lexical}, \cite{goldberg1990alternative}. From a robotic perspective, there are several examples of personality-adaptive verbal interactions \cite{andriella2022know}, \cite{Lim2022538}, \cite{Garello2020256}, \cite{Aly2016193}. In our research, we have carefully selected the language parameters based on previous studies \cite{nardelli2024ei} on dyadic conversations and literature analysis \cite{andriella2022know}, \cite{Lim2022538}, \cite{Garello2020256}, \cite{Aly2016193}. Parameters are directly integrated in the prompt for LLM, thus being incorporated in the language generation process, as described in Section 3.5. The defined parameters are listed below, in relation to the different personality traits.

\begin{itemize}
    \item Extroversion:  Verbose, Friendly, Talkative, Enthusiastic, Excited.
    \item Introversion:  Reserved, Quiet, Neutral.
    \item Agreeable: Cooperative, Friendly, Empathic, Forgiving, Reliable, Polite.
    \item  Disagreeable: Competitive, Aggressive, Provocative, Selfish, Rude.
    \item Conscientious: Scrupulous, Precise.
    \item  Unscrupulous: Thoughtless, Distracted, Lazy, Disordered.
\end{itemize}

Regarding arm gestures, previous literature correlates the amplitude and speed of gestures with extroversion \cite{craenen2018shaping} \cite{craenen2018we}. A broader perspective helps us understand how movement parameters are associated with personality traits \cite{delgado2022automatic}, identifying four parameters: velocity, speed, acceleration, and linearity of the trajectory. Extroversion is positively correlated with the speed and amplitude of gestures. Agreeableness is associated with acceleration and the linearity of the trajectory. Finally, Conscientiousness positively correlates with the directness of the trajectory \cite{delgado2022automatic}.

Next, we leverage the generative capabilities of a BERT language model to implement the Personality Generator as a flexible component that is pervasive across all the aforementioned behavioral parameters and is not dependent on a specific platform or context.

Specifically, we fine-tuned a BERT language model to input the defined personality (as outlined in Eq. (1)) and a general action to be executed, and to output the set of behavioral parameters required for that action. In order to have details about the procedure followed to generate the dataset and its association with the proposed CEA taxonomy, the machine learning problem addressed, the finetuning procedure and the performances obtained the reader can refer to our previous research \cite{nardelli2023software}.

The generalization capability inherent in language models enables us to create a component that functions as an abstract concept, universally applicable across different platforms, rather than merely representing a stereotype of a trait with a fixed set of behaviors.
%Please notice that, in case at least two coefficients of Eq.1 are not null (i.e., there are at least two personality dimensions that are not neutral), only one trait is extracted from the distribution of weights, influencing the action executing, with probability depending on the value of the coefficients. Accordingly, during the task, all traits are observed with no particular order depending on the distribution of weights. The result is a non-deterministic variation of personality throughout the interaction, providing a flexible approach. Finally, it is worth saying that in case actions are executed multiple times during the interaction (e.g., navigation, verbal interaction), parameters are generated only once, and they are kept constant for the whole interaction. This has been done to avoid having a weird outcome in case different personality dimensions have contrasting parameters. 

\subsection{Prospection and Episodic Memory}
\label{sec:prospection}

\begin{table*}[h]
\centering
\caption{Motivational Goals and associated verbal and movement actions. E, A, C are the three personality traits. L and H mean respectively high value and low value of the dimension of interest (e.g., LE means introversion and HE means extraversion).}
\label{table_example}
\begin{adjustbox}{max width=\textwidth}
\resizebox{18cm}{!}{%
\begin{tabular}{|p{1cm}|p{4cm}|p{6cm}|p{5cm}|}
\hline 
\rowcolor{lightgray} \textbf{Trait} & \textbf{Motivational Goals} & \textbf{Verbal Actions} & \textbf{Non-Verbal Actions}  \\ 
\hline
\textbf{HE} & Achievement, Excitement & 
Say an enthusiastic sentence, Tell a joke, Ask a question, Capture attention, Tell a personal story &
Make a visible movement horizontally, Make a visible movement vertically \\ 
\hline

\rowcolor{lightgray!30} \textbf{LE} & Detachment & 
Ask if you can be useful, Ask a reflective question, Say you prefer private conversation & 
Make a retracting movement, Hide the gripper behind the arm \\ 
\hline

\textbf{HA} & Compassion, Politeness & 
Express empathy, Give a compliment, Ask if you can help, Declare there's no reason to be angry &
Move closer to the human \\ 
\hline

\rowcolor{lightgray!30} \textbf{LA} & Selfishness & 
Make a contrastive statement, Express disapproval, Ask a provocative question, Insist that you are always right & 
Make a threatening move horizontally, Make a threatening move sagittally, Open and close the gripper to tease the human \\ 
\hline

\textbf{HC} & Industriousness, Orderliness & 
Remind the human to be focused, Offer guidance, Promote ethical behavior &
Make a gesture to keep the human’s attention on the task \\ 
\hline

\rowcolor{lightgray!30} \textbf{LC} & Unreliability & 
Distract with random questions, Make thoughtless remarks, Say inconsistent things &
Make a random movement, Wait some seconds \\ 
\hline
\end{tabular}%
}
\end{adjustbox}
\end{table*}

\begin{figure}[h]
\centering
\includegraphics[width=1.0\textwidth]{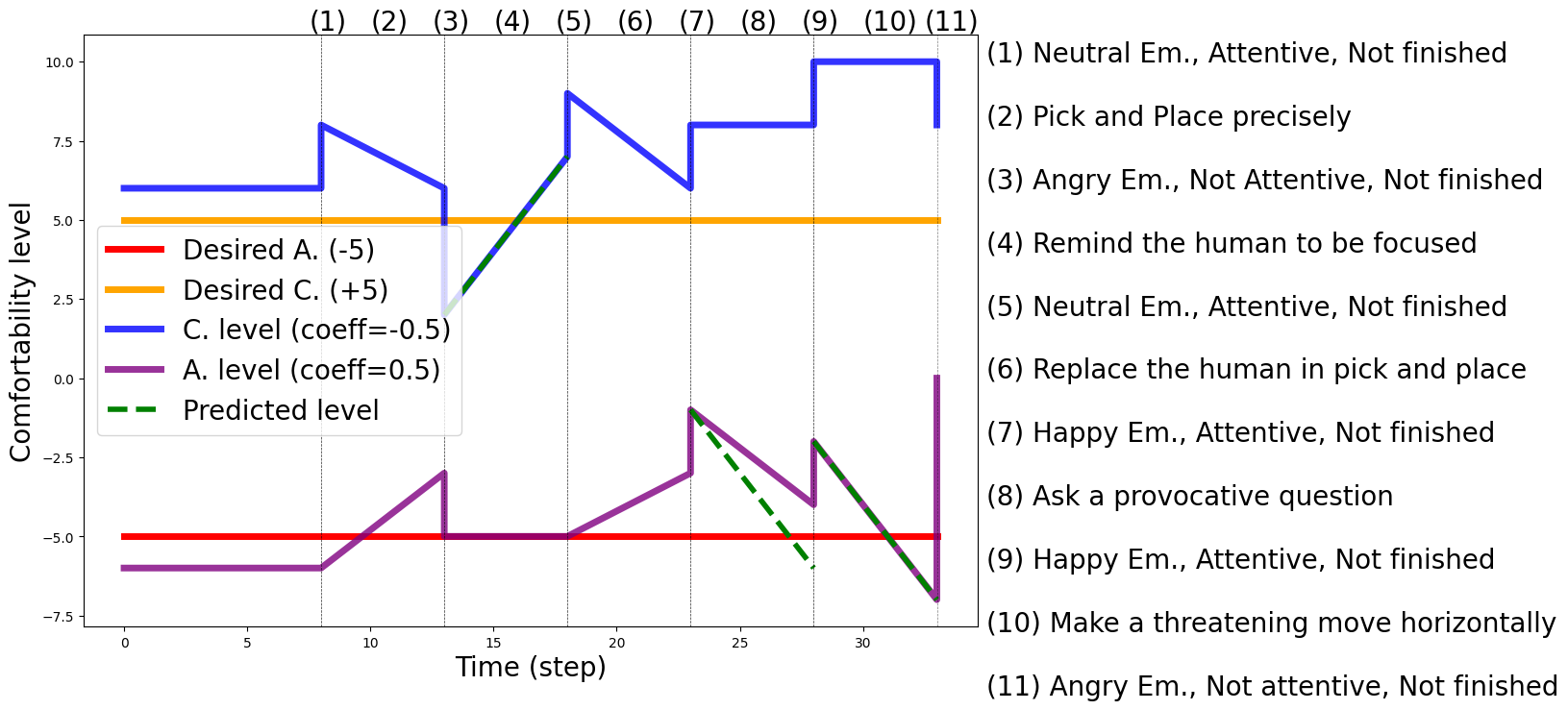}
\caption{The figure displays a snapshot of the plan retrieved from the Prospection process for a Conscientious and Disagreeable robot, showing the selection of actions, their assessments, and updates to the Episodic Memory. The reader may observe the conscientiousness level (in blue) and the agreeableness level (in purple) and their responses to different actions. The discrepancy between the green dashed lines (representing expected outcomes) and the solid blue/purple lines (depicting actual outcomes) highlights the difference between the anticipated rewards and the actual results.}\label{fig:Prospection}
\end{figure}

The human brain is so sophisticated that it allows individuals to satisfy their hedonic nature by planning sequences of actions in advance while maximizing comfortability \cite{gilbert2007prospection}. This cognitive capability, known as Prospection, relies on episodic memories to internally simulate future actions and the associated comfort. Specifically, it manages comfort through an allostatic control loop (predictive homeostasis) to select actions that optimize comfort \cite{vernon2014artificial}. Each personality trait is linked to specific brain areas that become more excited when certain motivational goals are achieved \cite{deyoung2010testing}. As a result, individuals will choose action sequences that best align with their own personalities to reach their goals \cite{vernon2014artificial} (Table 1, Motivational Goals). 
%In order to select the specific actions presented in Table I (Verbal Actions, Non-Verbal Actions) we started from existing psychological literature \cite{roccas2002big}, \cite{seligman2013navigating} which frame the relationship between values and OCEAN traits. %This relationship manifests itself in recurring, prospecting-driven behaviors associated with different personality traits. 
Starting from the motivational-goals associated behaviors, we inferred a set of actions that a Kinova Jaco2 arm should execute, associated to the identified personality traits, in a collaborative task (Table 1, Verbal Actions, Non-Verbal Actions). During this inference process, we considered psychological literature \cite{roccas2002big}, \cite{seligman2013navigating} about the relationship between personality, prospective, and action selection process and previous literature dealing with conveying social actions through robotics arms \cite{savery2020emotional}, \cite{la2022humans}. Please notice that, although the presented architecture is platform- and task-independent, the choice of actions as well as their implementation is of course related to the robotic platform used.

Cognitive abilities are not blind; they are responsive to the surrounding world. An individual's comfort is influenced not only by their own traits but also by their interactions with others and the environment they are in. For example, the perceived emotions of others can affect the comfort of a personality-cognitive agent \cite{nardelli2024ei}. An extroverted individual, for instance, tends to engage in actions that, at the same time, achieve their personality-dependent motivational goals and capture the attention of others, thereby inducing high-arousal emotions in those they interact with.

To effectively capture the ability to predict action tendencies based on specific personality traits, the proposed cognitive architecture models the Prospection module as a control system characterized by predictive control and feedback, known as allostasis \cite{vernon2014artificial}. We implemented the Prospection process using a FF planner \cite{hoffmann2001ff}, which allows the agent to simultaneously achieve tasks while maintaining personality-related comfort levels above a certain threshold. 
The system can predict the effect that action execution has on comfortability, nevertheless, this effect on the comfortability levels depends on several factors: in particular, the action's outcome (in term of emotions and attention of other individuals), and the past episodes of that specific action. The mechanism of expected reward and effectively retrieved feedback which control the Prospection and hence the action-selection process is presented in detail within this section.

The planner simultaneously controls three comfortability funtions which model the levels comfortability associated to  Extroversion, Agreeableness, and Conscientiousness. These are implemented through numerical fluents and allow us to simulate a variation of the agent's comfort through the planning procedure. These three levels do not affect each other and are concurrently present and managed separately by the planner. These levels must be regulated within the allostatic system to optimize in advance the agent's hedonic experience, ensuring their absolute value remain above a predefined comfort threshold. The threshold is set at a constant value at the start of the interaction (e.g., +Th or -Th, representing the positive and negative extremes of a specific trait, respectively). At a practical level, this is implemented by adding a subgoal in the planner that guarantees the absolute value of the comfortability functions (numeric fluents) to be kept above the absolute value of these thresholds.

Within the system, two types of actions are distinguished based on their impact on comfortability. Each type allows the agent to perform tasks while influencing its hedonic experience.

Firstly, \textit{standard actions within the plan} cause the absolute value of comfortability functions to decrease linearly over time, with a slope that is inversely proportional to the magnitude of the trait coordinate (i.e., the coefficient associated with that personality dimension (Eq. 1)). These actions refer to the activities that an agent has to execute to achieve a specific goal, and usually they do not
directly relate to the hedonic experience, not being connected to motivational goals. Consequently, the absolute value of comfortability decreases in time. The slope with which the comfortability function decreases due to the execution of standard actions models the preponderance with which a personality trait manifests itself (i.e., the higher the weight of the personality trait (Eq. 1), the higher the slope).  

Occasionally, the system offers alternative paths to achieve the same goal, known as complementary actions. These complementary actions allow the agent to achieve the same goal but have a different impact on comfortability, since they may involve social or environmental interactions which may be more or less preferred depending on the personality. This impact has been implemented by adding or subtracting an offset to the linear variation of the specific trait-related comfortability functions interested by the complementary actions (Fig. \ref{fig:Prospection}, actions 2 and 6). 
%Each action (complementary or not) has its own effect on all the three dimensions of personality, in case of a complementary action the offset is applied only to the identified trait whereas the absolute value of the remaining two comfortability fluents still decrease linearly in time. 

 %Once a \textit{standard action within the plan} is executed, the feedback obtained—in terms of comfort variation—matches the simulated outcome.
  
As a result, the system aims to trigger a sequence of actions that maximizes comfortability without adhering to a specific path, anticipating future actions, exploring alternative options, and avoiding uncomfortable solutions whenever possible. Indeed, when the agent needs to pick and place an item on a board, possible actions include \textit{Pick and place precisely} or \textit{Pick and place incorrectly}. The first action will add a positive (always in term of absolute values) offset on the comfortability value related to the Consciousness dimension, while the second one will have a positive effect on the comfortability value related to Unconsciousness. Hence, depending on the personality of the robot (i.e., from an implementation perspective, the comfortability function and related threshold considered by the planner) different sequences of actions could be selected to achieve the same goal. To further clarify the concept, a disagreeable personality might prefer \textit{Replace the human in pick and place} as it aligns with their more arrogant nature. Alternatively, a rude agent might choose the sequence: \textit{Tell to the human that it is its turn.} followed by \textit{Make a threatening move horizontally}. Both approaches allow the system to accomplish the task and meet motivational goals.

For the definition of standard and complementary actions, once again we inferred from the psychological literature \cite{roccas2002big}, \cite{seligman2013navigating} which task-specific actions do not enable the achievement of motivational goals, and which actions instead affect the hedonic experience of a specific personality pole.

%In contrast, when the robot has to advise the human that is its turn, alternatives include \textit{Tell the human it is their turn}, \textit{Notify the human that it is their turn with a wave sound}, \textit{Replace the human in picking and placing without saying anything}, or \textit{Replace the human and advise them}.
%In the first condition of the experiment, where the robot can only move and not communicate verbally, only actions that do not involve verbal communication can be triggered. These actions might reduce the comfort level of an extroverted function, causing a highly extroverted robot to perform more noticeable movements to attract attention. In the second condition, where the robot can also communicate verbally, a highly extroverted robot is more likely to \textit{Tell the human it is their turn}. However, it can also \textit{Advise the human it is their turn with a wave sound} and then \textit{Say an enthusiastic sentence} to increase comfortability.
%Given the nature of the planner, different sequences of actions may be selected to achieve the given goal, on the condition that the comfortability levels always meet the given threshold.

%Prospection process also models the dominance of personality traits. In a collaborative task, a more dominant individual may start the task and take additional actions compared to others. An individual who likes to show off or is selfish (high extroversion, low agreeableness) might be rewarded for starting the task first, though they are not compelled to do so.

Secondly, there are \textit{motivational goal actions} (Table 1 and Fig \ref{fig:Prospection}, actions 4, 8, 10). These actions are initiated when the system predicts that the absolute value of the comfortability functions related to the implemented personality will fall below the given threshold (e.g., with reference to Fig. \ref{fig:Prospection}, action 4 is triggered because the Conscientiousness level falls below the threshold, actions 8 and 10 because the (negative) value of Disagreableness goes above the related threshold). In such cases, the system needs to enhance the hedonic experience by executing motivational goal actions. These actions do not contribute directly to achieving the primary goal but are aimed at maintaining or improving comfortability. 

As previously mentioned, the slope with which the absolute value of a comfortability function decreases indicates the extent to which the related personality trait is expressed. Therefore, a higher angular coefficient, reflecting a stronger manifestation of that personality dimension, leads to a greater frequency of trait-related motivational goal actions being triggered. 

When a motivational goal action needs to be executed, the system consults the Episodic Memory to select the most rewarding action (i.e., the ones that mostly increase, in terms of absolute value, the comfortability level). %Table I lists the possible actions associated with each personality pole. In the first condition, the robot can only trigger movement actions. In the second condition, where the robot can also engage in verbal interactions, both movement and speaking actions are available. 
In order to execute contextualized motivational goal actions, we represented the possible world state as a bit array, which encodes the user's detected emotion and attention. The reward associated with each encoded world state and set of possible actions is initialized a priori and depends on two factors: one is based on the appropriateness of the action in that specific context, which remains fixed over time; the other depends on the expected outcome of the action on the environment, which varies over time based on the real outcome of the action execution.

Given the robotic personality, when a motivational goal action needs to be executed, the current emotion felt by the user and his attentional state are detected by the perception blocks (Fig. \ref{fig:Prospection}, steps 1, 3, 5, 7, 9, and 11). In this context, the most appropriate action is selected by using the different rewards of actions associated with the actual world state to compute a weighted random probability, as this process is likely to provide the most promising action to yield a significant reward. The randomness of the selection process ensures that the system explores all available actions rather than focusing on just one. After executing the action, the perception is reassessed. If it aligns with the predicted emotion (Fig. \ref{fig:Prospection}, steps 3, 4, 5 and 9, 10, 11), the obtained reward is increased; otherwise (Fig. \ref{fig:Prospection}, steps 7, 8, 9), it is decreased.

The Episodic Memory encodes the link between the environmental state, the chosen action, and the obtained reward (in terms of comfortability). Consequently, whenever there is a mismatch between the expected and actual rewards, the probability of selecting the action that caused the diminished reward decreases. For instance, a disagreeable agent, upon perceiving someone happy, might choose to \textit{ask a provocative question} to provoke anger. If, after executing the action, the person keeps smiling, the obtained reward will be diminished and stored in the Episodic Memory.

To sum up, with the described methodology, each time an action needs to be executed, it is not chosen entirely at random (unlike in the previous implementation \cite{nardelli2024personality}), but rather it depends on the current environmental state, the expected outcome of the action, and prior experiences with that action.

Through the integration of Episodic Memory and Prospection, the cognitive architecture can tailor interactions based on the robotic personality's internal objectives and the individual user's reactions to stimuli, creating an internal experience customized to these factors. %For example, an agreeable personality would strengthen its repertoire of actions aimed at eliciting happiness in a specific user.

Finally, values of comfortability may also vary asynchronously, independently from whether an action is performed, and only in relation to the observation done through the interaction with the environment. This mechanism can potentially integrate any perception stimuli. In this implementation, the perception of the user's emotional and attentional states can positively or negatively influence the trait-related comfortability. Indeed, this perception can directly alter the levels of the comfortability functions, reflecting the sensitivity of that specific personality trait to the perceived stimuli \cite{pease2015personality}, \cite{deyoung2020personality},\cite{hughes2020personality}. This variation of comfortability can trigger the execution of \textit{motivational goal actions} without any explicit need to achieve a practical goal, thus giving the impression of an actually proactive agent in his actions.

%It is important to notice that the implemented process allow to capture the suboptimal, unrepresentative, essentialized, decontextualized nature of Prospection as highlighted by research \cite{gilbert2007prospection}.

\subsection{Execution Components}
The combination of Personality Generator and Prospection allows the cognitive architecture to implement personality as a psychological component, influencing how actions are executed and which actions are chosen, thereby shaping the overall interaction. Although the Personality Generator and the predicted behavioral parameters, the memory capabilities, and the Prospection process have been designed in order to reflect task- and platform- independence, the specific execution blocks directly depend on the robot used and the task where it is involved.
To design and implement the execution components within the Kinova Jaco2 arm we considered the possible set of actions that the robot can effectively execute.
Dealing with movement, a manipulator does not allows us to implement actions that involve head movements, navigation, and gaze \cite{nardelli2023software}, \cite{nardelli2024personality}. On the other hand, we exploited the dexterity of the Kinova Jaco2 to pick and place objects and perform communicative actions.
Dealing with verbal communication we effortlessly incorporate a microphone within the robotic system that allows the robot to communicate directly through speech.
Consequently, the action components, differently from our previous research \cite{nardelli2023software}, \cite{nardelli2024personality} where all the personality influenced parameters were effectively applicable, retrieve from the Personality Generator only speech (Volume, Language) and gesture (Amplitude of Trajectory, Velocity, Acceleration, and Straightness of trajectory) parameters (Section \ref{sec:psycho}) which code the manipulator movement and influence the sentence and voice generation processes. 
To validate the proposed cognitive architecture we implemented a collaborative task between the user and the robot. Specifically, the robot should collaborate with a human to fill a chessboard with LEGO blocks of different colors, since this task was well suited to explore different actions, with their choice being potentially affected by personality. The task is explained in detail in the Experimental Set-Up section (Section \ref{sec:es}).

\subsubsection{Speech Interface}

The Speech Interface component is responsible for generating the sentences that the robot vocalizes and producing the corresponding neural voice. To achieve this, we leverage GPT-4o \cite{chatgpt} for sentence generation and Microsoft Azure \cite{azure} to synthesize audio streams using a neural voice, which is then played through the robot’s microphone. Additionally, the Speech Interface can play wavesound effects upon request.

When a sentence needs to be generated, GPT-4o \cite{chatgpt} takes as input the prompt, the current input, and the history of previously generated sentences. The prompt for generating sentences is as follows:

\textit{You are the speech generator of a Kinova robotic arm. The robot is playing a collaborative game with a user, aiming to fill a chessboard with colored LEGO blocks. You need to generate sentences to verbally interact with the user within the} \texttt{text} \textit{field.}

\textit{The input provides several details:} 
\begin{itemize} \item \textit{the user's emotion, indicated in the} \texttt{emotion} \textit{field} 
\item \textit{the user's attentive state, indicated in the} \texttt{attention} \textit{field} 
\item \textit{the robot’s personality, specified in the} \texttt{personality} \textit{field, which influences the response style} 
\item \textit{the language style parameter, specified in the} \texttt{language style} \textit{field, guiding the tone and formality}
\item \textit{the (verbal) action type, used to generate an appropriate response, indicated in the} \texttt{action} \textit{field} \end{itemize}

The actual input is composed of the following information.
\begin{itemize}
    \item The user's \texttt{emotion} and \texttt{attention} are predicted by perception modules.
    \item The robot's \texttt{personality} and \texttt{language style} parameters are provided by the Personality Generator.
    \item The set of possible actions, to report within the \texttt{action} field, are determined by the Prospection component. These include the verbal actions required to performing the task (\textit{Informing the human it is their turn} and \textit{Advising the human that the robot has taken their turn}) and the motivational actions (Table 1, Verbal Actions column).
\end{itemize}

Once the sentence is generated, the Speech Interface requests Microsoft Azure \cite{azure} to synthesize the audio stream, which is then played at a volume that corresponds to the predicted parameter.

\subsubsection{Gesture Interface}
\label{sec:gesture}
\begin{figure}[h!]
\centering
\begin{minipage}{.45\textwidth}
\centering
\includegraphics[width=\linewidth]{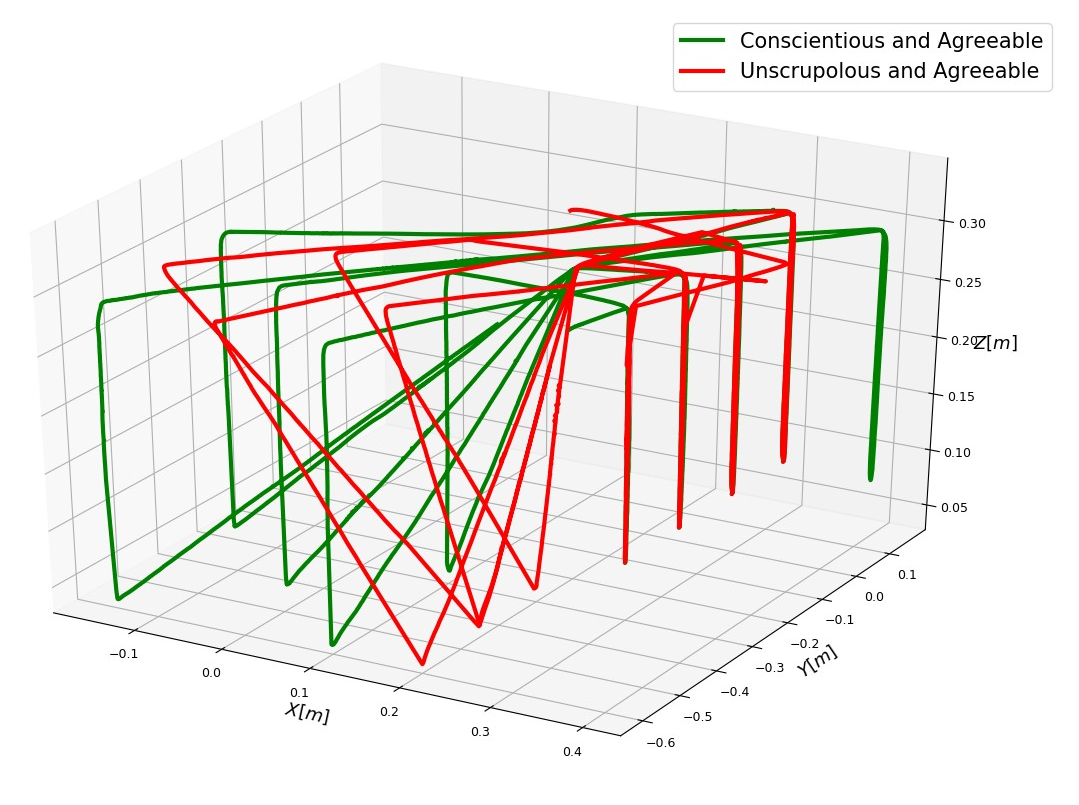}

\end{minipage}\hfill
\begin{minipage}{.45\textwidth}
\centering
\includegraphics[width=\linewidth]{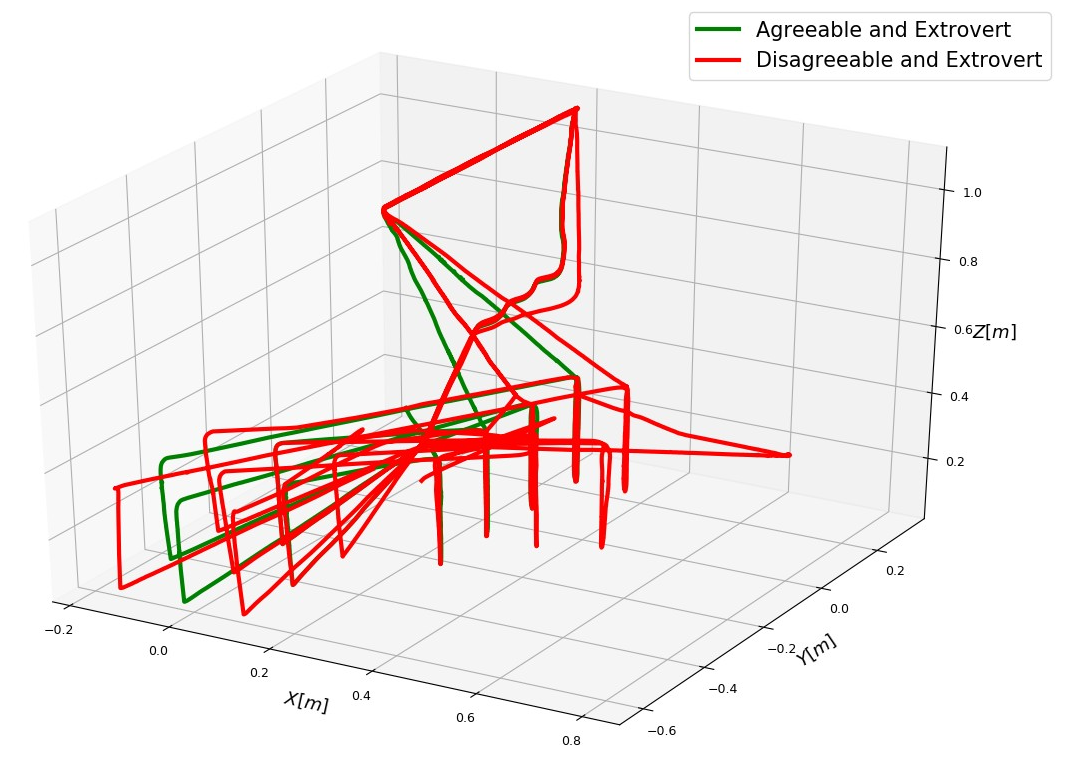}

\end{minipage}\hfill
\begin{minipage}{.45\textwidth}
\centering
\includegraphics[width=\linewidth]{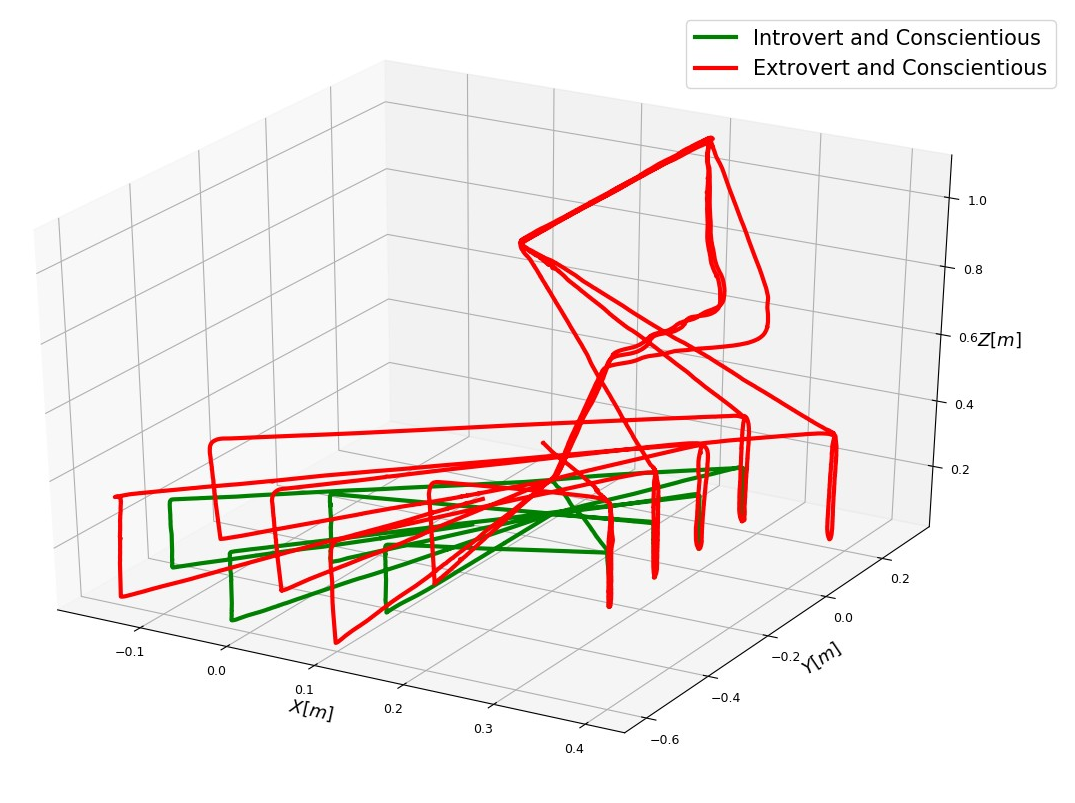}

\end{minipage}
\caption{The three plots show the three-dimensional trajectory of the robot involved in the implemented task. The plots highlight the differences between the two extremes of each trait, respectively Conscientiousness, Agreeableness, and Extroversion. In each plot, the trajectory of the negative pole is displayed in green, and the positive pole in red. }
\label{fig:traject}
\end{figure}

\begin{figure}[h!]
\centering
\includegraphics[width=0.9\textwidth]{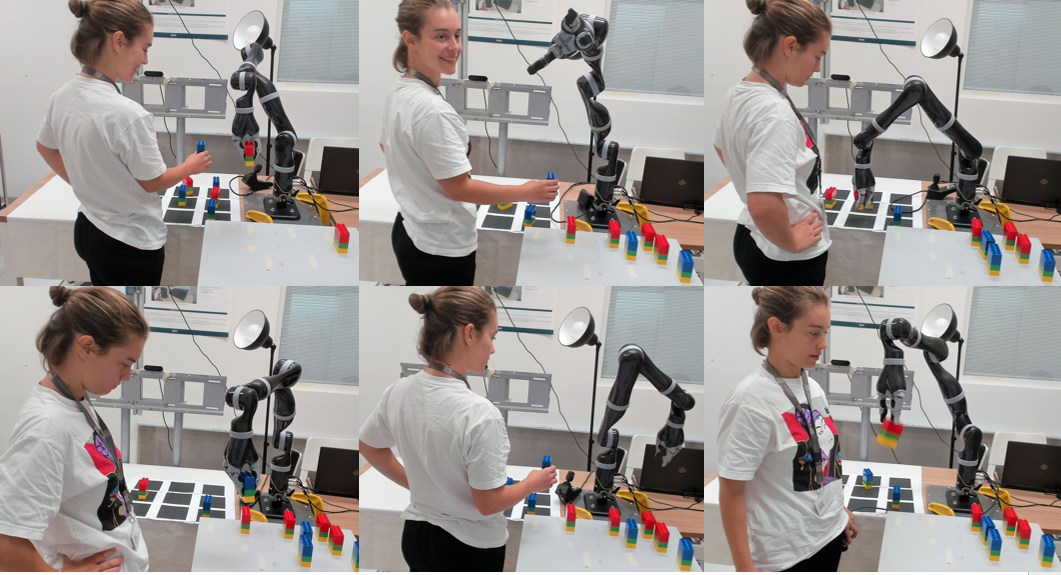}
\caption{Effect of a specific personality pole on the interaction. From left to right.
Top: Agreeable, Extrovert, Conscientious personality.
Bottom: Disagreeable, Introvert, Unscrupolous personality.}
\label{fig:interaction}
\end{figure}

To implement the Gesture interface, we utilized the Kinova Jaco2 SDK. The personality-dependent parameters were embedded within the robot's movements and are predicted dynamically whenever an action involving arm movement needs to be executed.

The Velocity and Acceleration parameters were incorporated by configuring the corresponding settings within the SDK. The High Velocity and High Acceleration levels were associated with the robot's supported maximum velocity and acceleration, the Middle values with 75\% of the maximum values, and the Slow values with 50\%.

The directness of the trajectory was implemented by introducing waypoints that deviate slightly from the direct path, creating a less linear trajectory. On the other hand, the Amplitude of the trajectory was implemented by appropriately defining the z-coordinates of the pre-grasp, post-grasp, pre-release, and post-release waypoint positions.

Regarding movements and actions, we implemented the set of actions required to perform the collaborative task (Figure \ref{fig:interaction}): \textit{Pick and place precisely}, \textit{Pick and place wrongly}, \textit{Replace the human in pick and place}. Additionally, we included motivational goals action (Table 1, Non-Verbal Actions column). In order to convey messages through physical gestures we took into account the consideration done by related literature about the implementation of basic emotions in robotic arms \cite{savery2020emotional}, \cite{la2022humans}. 

Figure \ref{fig:traject} shows a three-dimensional plot of the trajectory recorded during the user study, highlighting the differences between opposite poles of a specific trait. High extroversion is evident in the broader amplitude of gestures used to perform pick-and-place actions. Additionally, the visible movements of the sociable robot to capture attention can be clearly observed.

The difference between an agreeable and disagreeable robot is noticeable in two main aspects. The concept of dominance and the preference of a disagreeable robot to replace human actions are reflected in a higher number of actions performed by a rude robot. It is also evident that the agreeable robot follows a more direct trajectory compared to the disagreeable one. This distinction is also apparent when comparing high and low conscientiousness. 

% Finally, it is noticeable how the distracted robot makes errors in placing the block. These differences can be considered the bone of effective interaction. They shape the interaction at a fundamental level by influencing human behavior (Figure \ref{fig:interaction}). The agreeable robot works in synergy with the participant, whereas the disagreeable robot tends to replace them. The extroverted robot makes noticeable movements to capture the human's attention, while the introverted robot retracts to avoid drawing attention. Finally, the conscientious robot performs the actions correctly, while the distracted robot makes mistakes in completing the task.

\section{Study Design} 
\label{sec:es}

To test the capabilities of the proposed cognitive architecture for implementing the CEA taxonomy personality in a robotic arm, an experiment involving a Jaco2 Kinova arm and a collaborative game has been set up. 
\subsection{Research Questions and Hypothesis}
We organized the user study to address two research questions:

\begin{itemize}
\item \textit{Is it possible to implement robot personality through the CEA cognitive architecture proposed in \cite{nardelli2024ei} on a robotic arm? Are these personalities perceivable by humans?}
\item  \textit{What impact has verbal interaction on the perception of personality?}
\end{itemize}

The evaluation is based on two hypotheses:

\begin{itemize}
\item \textit{Hypothesis I}: The cognitive architecture can generate distinguishable personality traits in a Jaco2 Kinova arm, which communicates personality solely through the quality of its movements and choice of actions.
\item \textit{Hypothesis II}: The perception of personality traits in this context is enhanced by verbal communication.
\end{itemize}

We believe that differences in personalities emerging from our software architecture can shape the interaction at a fundamental level by influencing human behavior (Figure \ref{fig:interaction}). It is expected that the agreeable robot may work in synergy with the participant, whereas the disagreeable robot will tend to replace them. The extroverted robot should make noticeable movements to capture the human's attention, while the introverted robot will retract to avoid drawing attention. Finally, the conscientious robot may perform the actions correctly, while the distracted robot will likely make mistakes in completing the task.

\subsection{Game-adapted Cognitive Architecture}
In the experiment, the robot is integrated with cameras and a microphone to enable it to interact with the user and the environment. During the game, the user and the robot have to fill a 3x3 chessboard with red and blue Lego blocks (Figure \ref{fig:interaction}). The red blocks are assigned to the robot and the blue ones to the human. At the start, all the blocks are placed in specific locations. The robot and the human are required to fill the board so that the same colours are never adjacent. The game is based on the mathematical theorem \textit{Four-Colour Theorem}\cite{robertson1997four}, which states that on a flat surface divided into interconnected regions, such as a political map, it is possible to color each region using only four colors, ensuring that no two adjacent regions share the same color. This enables the implementation of a neutral collaborative task governed by simple, well-defined rules, providing a set of possible actions through which personality can be effectively expressed.

The overall interaction is orchestrated by the robot and its priori-assigned personality, which determines who initiates the interaction and when it is the human's turn, communicating this either verbally or with a sound cue. Participants are instructed to pick a blue block and place it on the chessboard whenever advised. The human and the robot are not required to follow a specific pattern in completing the task; the only constraint is to fill the chessboard in an alternating sequence (e.g., blue, red, blue, ...).

Moreover, to compare the impact of verbal interaction within the perception of personality we compared two conditions: in the first condition the robot can only move and communicate through sounds (\textit{Not-Speaking} condition), whereas in the second one, it can also communicate through language (\textit{Speaking} condition).

In order to implement this behavior we equipped the framework (Section \ref{sec:mm}, Figure \ref{fig:sofar}) of a Board Perception block able of perceiving the actual state of the chessboard, by means of the OpenCV library \cite{opencv1} to detect the chessboard and LEGO blocks. The Board Perception block communicates to a Game Player component which chessboard's areas are free and which are occupied by a blue or red LEGO block. The Game Player, once the first block is placed upon the chessboard, infers the final configuration (since only two \textit{correct} configurations are available, one with the blue block in the center, one with the red block in the center). The Action Dispatcher queries the Game Player to get informed about the new move (i.e., where the next block should be placed).

The Prospection module has also been tailored to this specific experiment: in the \textit{Not-Speaking} condition of the experiment, where the robot does not communicate verbally, the domain only includes actions that do not involve verbal communication. The consequent choice of actions (e.g., \textit{Notify the human that it is their turn with a wave sound}) sharply reduces the comfort level of the comfortability value associated to Extroversion, causing a highly extroverted robot to perform more noticeable movements to attract attention. In the \textit{Speaking} condition, where the robot can also communicate verbally, a highly extroverted robot is more likely to verbally notice the user that it is their turn to move. Dealing with \textit{motivational goal actions} (Table 1) in the \textit{Not-Speaking} condition, the robot can only trigger movement actions. In the \textit{Speaking} condition, where the robot can also engage in verbal interactions, both movement and speaking actions are available. 

Focusing on the execution blocks the Speech Interface in the \textit{Not-Speaking} condition can only reproduce wavesound, while in the \textit{Speaking} condition it also manages the vocal interaction. Dealing with the Gesture Interface, please notice that all tasks including the interaction with LEGO blocks have been implemented by knowing a priori their initial positions, as well as the absolute positions of the different cells of the chessboard, and by detecting the positions of the LEGO blocks within the chessboard.

\subsection{Experimental Set-up}
Tests have been conducted using an experimental between-subjects design with two conditions (\textit{Non-Speaking} vs. \textit{Speaking}). We have used the cognitive architecture's capability to manage multiple traits simultaneously, combining each of the two poles of one specific trait with the four poles of the remaining two traits. This resulted in 12 personalities to test for each condition. Consequently, we opted for a fractional factorial partial within-subjects design (robotic personality), where each participant engaged in three sessions, experiencing three combinations of personality traits that are shuffled in advance to balance the 12 personality traits and minimize the order effect \cite{collins2009design}.

We recruited a total of 48 participants: 24 subjects interacting with the robot in the \textit{Non-Speaking} condition, and 24 interacting with the robot in the \textit{Speaking}condition. Participants were recruited through an online process. All participants provided written informed consent before starting the experiment.

%Upon entering the laboratory, each participant is asked to sign the informed consent form. The experimenter then explains the specific task they have to complete with the robot. 
As mentioned, the experiment consisted of three sessions, during which the robot displayed different combinations of traits, so that each participant during the experiment interacted with all the 6 personality poles. Consequently, for each condition (\textit{Non-Speaking} vs. \textit{Speaking}) we collected a total of 72 sessions, and each of the 12 personalities was tested 6 times, with each personality pole being tested 24 times. Each session ended when the participant had completed the chessboard. 

\subsection{Measurement}
To validate the presented hypotheses, at the end of each interaction, participants were asked to provide demographic information and complete a 5-point Likert scale questionnaire about their impression of the robot's behavior during the just-concluded session. The questionnaire administered to users was the Italian-validated version of the 10-item Big Five Inventory \cite{guido2015italian}, suitably adapted to the third person to assess perceived robotic personality (Hypotheses I and II). The questionnaires was administered online through the open-source SoSci Survey platform \cite{soscisurvey1}. The items of the questionnaire were randomly shuffled to prevent any order effect, and an attention-check item (\textit{Answer 5 to this question to demonstrate your attention}) was included to ensure that participants were focused while completing the questionnaire.

Additionally, participants were asked to answer the open-ended question, \textit{"Write your impression of interacting with the robot"}. This allowed us to obtain a more nuanced evaluation of the perception of the robotic personality and the resulting feelings of the participants.

\section{Results And Discussion}
To address our research questions and determine whether it is possible to implement a set of personalities in a non-humanoid robot perceivable by humans through the proposed cognitive architecture, we adopted a two-fold strategy. On one hand, we analyzed the responses from the BFI 10-item questionnaire; before performing the statistical analysis of the questionnaire results, we ensured the reliability of all scales used in the implemented tests. This was done by calculating Cronbach's alpha \cite{cronbach1951coefficient}, a widely recognized measure of internal consistency. All scales were confirmed to have a coefficient greater than 0.7, indicating acceptable reliability.
On the other hand, we systematically analyzed the responses to the open-ended question.

\subsection{Perception of Personality - Questionnaire}

\begin{figure}
\centering
\begin{minipage}{.45\textwidth}
\centering
\includegraphics[width=\linewidth]{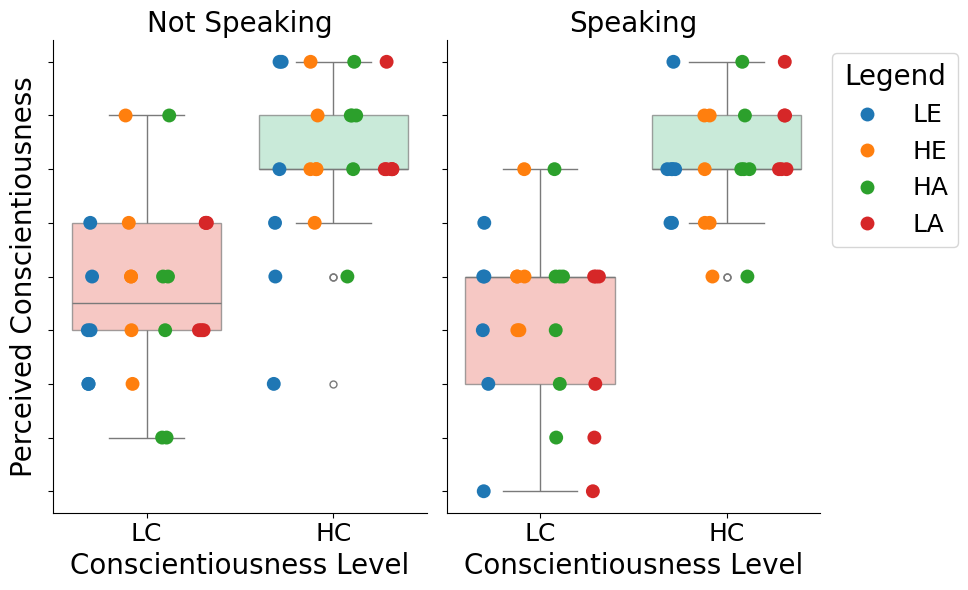}
\end{minipage}\hspace{-0.2cm}  % Adjust the negative space here
\begin{minipage}{.45\textwidth}
\centering
\includegraphics[width=\linewidth]{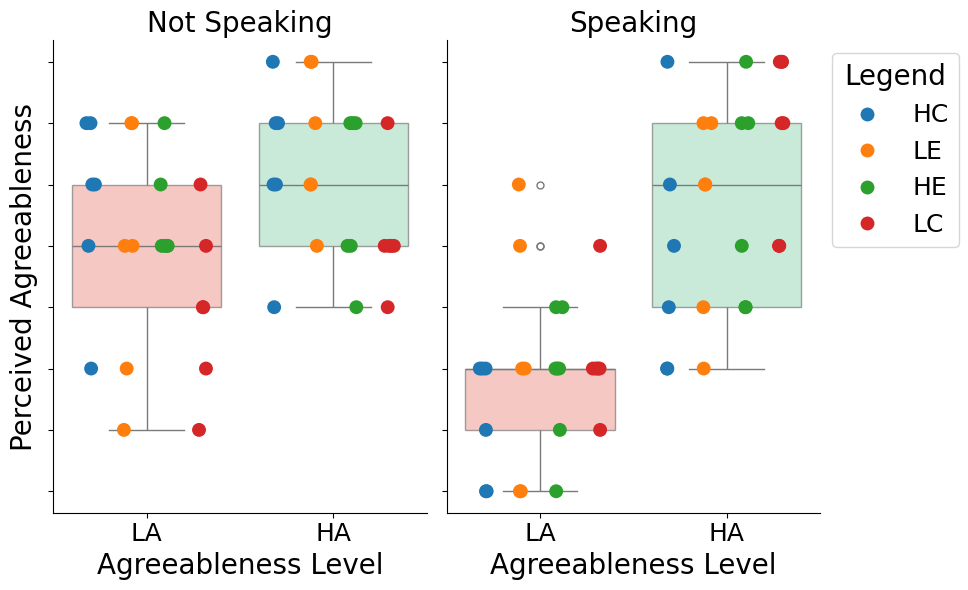}
\end{minipage}\hspace{-0.2cm}  % Adjust the negative space here
\begin{minipage}{.45\textwidth}
\centering
\includegraphics[width=\linewidth]{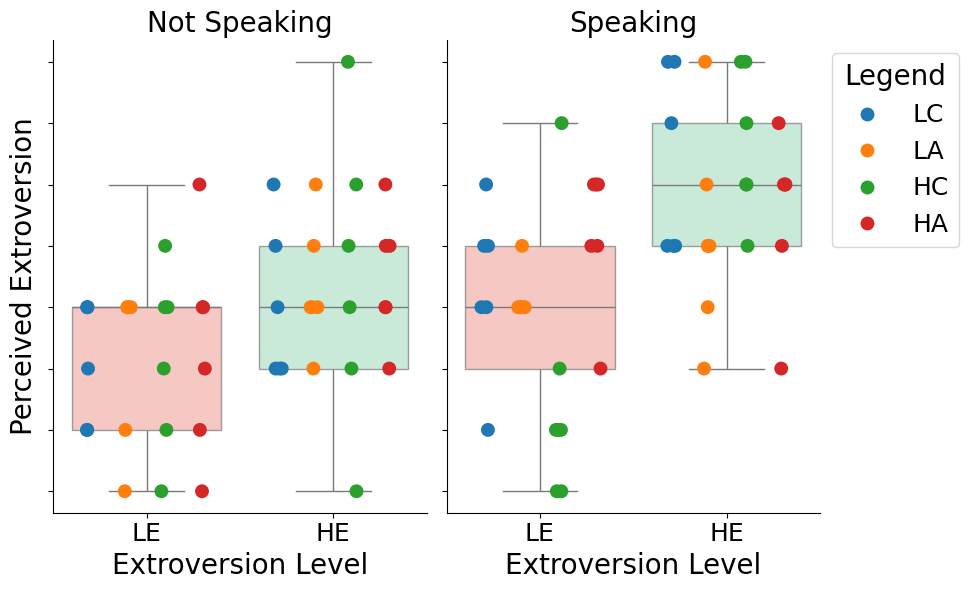}
\end{minipage}
\caption{Each plot corresponds to one trait of the CEA taxonomy and represents how that trait was perceived by subjects during the interaction (measured in range 1-5 according to the Likert scale adopted to administer the questionnaire). The box represents the quartiles of the dataset considered, while the whiskers indicate the remaining distribution. The scatterplot shows the perception of the pole of interest of each participant. The hue of the scatterplot indicates how the specific pole of interest is combined with the poles of the remaining two traits in the taxonomy. The plots also distinguish between the two conditions considered: Speaking and Not Speaking.}
\label{fig:boxplot}
\end{figure}

\begin{table*}[h]
\centering
\caption{Results of the Mann-Whitney U test \cite{nachar2008mann} to investigate if opposite personality poles are perceived as distinct along the corresponding trait. The considered test Hypothesis is $G1 \neq G2$.}
\label{table_example}
\begin{adjustbox}{max width=\textwidth}
\resizebox{18cm}{!}{%
\begin{tabular}{|p{4cm} p{4cm} p{2cm} p{4cm} p{4cm} p{1cm} p{1cm}|}
\hline 
\rowcolor{lightgray}\textbf{Independent Variables G1, G2}&\parbox{5cm}{ \textbf{Dependent Variable}}& \textbf{Condition} & \textbf{G1: Mean, Mediand, STD, N} & \textbf{G2: Mean, Mediand, STD, N} & \textbf{U } & \textbf{p value} \\

\parbox{3cm}{\textbf{LE-HE}} & \parbox{3cm}{Extroversion} & \parbox{3cm}{\textit{Not-Speaking}}  & \parbox{3cm}{2.60, 3, 0.659, 24} & \parbox{3cm}{3.17, 3, 0.732, 24} & \parbox{3cm}{172}& \parbox{3cm}{0.013}\\ 

\parbox{3cm}{\textbf{LE-HE}} & \parbox{3cm}{Extroversion} & \parbox{3cm}{\textit{Speaking}}  & \parbox{3cm}{3, 3, 0.829, 24} & \parbox{3cm}{3.94, 4, 0.740, 24} & \parbox{3cm}{129}& \parbox{3cm}{$<$0.001}\\

\parbox{3cm}{\textbf{LA-HA}} & \parbox{3cm}{Agreeableness} & \parbox{3cm}{\textit{Not- Speaking}}  & \parbox{3cm}{3, 3, 0.78, 24} & \parbox{3cm}{3.50, 3.50, 0.643, 24} & \parbox{3cm}{191}& \parbox{3cm}{0.04}\\ 

\parbox{3cm}{\textbf{LA-HA}} & \parbox{3cm}{Agreeableness} & \parbox{3cm}{\textit{Speaking}}  & \parbox{3cm}{1.92, 2, 0.656, 24} & \parbox{3cm}{3.40, 3.50, 0.866, 24} & \parbox{3cm}{58.5}& \parbox{3cm}{$<$0.001}\\ 

\parbox{3cm}{\textbf{LC-HC}} & \parbox{3cm}{Conscientiousness} & \parbox{3cm}{\textit{Not-Speaking}}  & \parbox{3cm}{2.83, 2.75, 0.789, 24} & \parbox{3cm}{4.08, 4, 0.732, 24} & \parbox{3cm}{72.2}& \parbox{3cm}{$<$0.001}\\

\parbox{3cm}{\textbf{LC-HC}} & \parbox{3cm}{Conscientiousness} & \parbox{3cm}{\textit{Speaking}}  & \parbox{3cm}{2.62, 3, 0.794, 24} & \parbox{3cm}{4.06, 4, 0.546, 24} & \parbox{3cm}{38}& \parbox{3cm}{$<$0.001}  \\
\hline
\end{tabular}%
}
\end{adjustbox}
\end{table*}

To evaluate the cognitive architecture's efficacy in generating human-perceivable robotic personalities, we analyzed responses from the BFI-10 questionnaire using the Mann-Whitney U test \cite{nachar2008mann}. In this analysis, the contrasting personality poles of the same trait exhibited by the agent are considered independent variables, while the mean scores of trait-related items on the questionnaire serve as the dependent variable. Our goal is to determine whether variations in each trait are perceptible to participants, specifically whether users could distinguish between introverted and extroverted, agreeable and disagreeable, or conscientious and distracted agents (\textit{Hypothesis I}). Table 2 shows the complete statistics of this analysis.

The results of the experiment in the \textit{Not-Speaking} condition indicate that variations in personalities across the three traits were accurately perceived by participants. They were able to discern whether the Kinova arm displayed positive or negative traits in a statistically significant manner (Agreeableness: $p=0.04$, Extroversion: $p=0.013$, Conscientiousness: $p<0.001$). Similarly, in the \textit{Speaking} condition, variations in personality along the three axes were also consistently perceived by users (Agreeableness: $p<0.001$, Extroversion: $p<0.001$, Conscientiousness: $p<0.001$). 

The box plot shown in Figure \ref{fig:boxplot} shows the results of the perception of the three considered traits in the \textit{Speaking} and \textit{Not-Speaking} conditions. It highlights that participants accurately perceived variations in personality along the three traits in both conditions (confirming \textit{Hypothesis I}). While we anticipated the system’s complete capability to generate perceivable personality traits, consistent with previously obtained results from the cognitive architecture \cite{nardelli2023software}, \cite{nardelli2024personality}, \cite{nardelli2024ei}, the findings are particularly notable in the \textit{Not-Speaking} condition. There results demonstrate the effectiveness of the system in producing perceivable personality traits in a distinct manner, even in the absence of verbal interaction. This outcome validates the choice of parameters and actions, including any assumptions made during their definition.

\begin{table*}[h]
\centering
\caption{Results of the Mann-Whitney U test \cite{nachar2008mann} to investigate if the language has an impact on the perception of personality in one single personality pole. Within the table, the two conditions \textit{Not-Speaking} and \textit{Speaking} are respectively abbreviated with NS and S. The considered test Hypothesis is $G1 \neq G2$.}
\label{table_example}
\begin{adjustbox}{max width=\textwidth}
\resizebox{16cm}{!}{%
\begin{tabular}{|p{4cm} p{4cm} p{4cm} p{4cm} p{1cm} p{1cm}|}
\hline 
\rowcolor{lightgray}\textbf{Independent Variables G1, G2}&\parbox{5cm}{ \textbf{Dependent Variable}} & \textbf{G1: Mean, Mediand, STD, N} & \textbf{G2: Mean, Mediand, STD, N} & \textbf{U } & \textbf{p value} \\

\parbox{3cm}{\textbf{LE(NS) - LE(S)}} & \parbox{3cm}{Extroversion}& \parbox{3cm}{2.6, 3, 0.659, 24}  & \parbox{3cm}{3, 3, 0.29, 24} & \parbox{3cm}{210}& \parbox{3cm}{0.065}\\ 

\parbox{3cm}{\textbf{HE(NS) - HE(S)}} & \parbox{3cm}{Extroversion}  & \parbox{3cm}{3.17, 3, 0.732, 24} & \parbox{3cm}{3.94, 4, 0.74, 24} & \parbox{3cm}{137}& \parbox{3cm}{$<$0.001}\\ 

\parbox{3cm}{\textbf{LA(NS) - LA(S)}} & \parbox{3cm}{Agreeableness}  & \parbox{3cm}{3, 3, 0.78, 24}  &  \parbox{3cm}{1.92, 2, 0.656, 24}  & \parbox{3cm}{96}& \parbox{3cm}{$<$0.001}\\ 

\parbox{3cm}{\textbf{HA(NS) - HA(S)}} & \parbox{3cm}{Agreeableness}  & \parbox{3cm}{3.50, 3.50, 0.643, 24}  &  \parbox{3cm}{3.40, 3.50, 0.866, 24}  & \parbox{3cm}{286}& \parbox{3cm}{0.775}\\ 

\parbox{3cm}{\textbf{LC(NS) - LC(S)}} & \parbox{3cm}{Conscientiousness}  & \parbox{3cm}{2.83, 2.75, 0.789, 24}  & \parbox{3cm}{2.62, 3, 0.794, 24}  & \parbox{3cm}{270}& \parbox{3cm}{0.538}\\ 

\parbox{3cm}{\textbf{HC(NS) - HC(S)}} & \parbox{3cm}{Conscientiousness}  & \parbox{3cm}{4.08, 4, 0.732, 24}  &  \parbox{3cm}{4.06, 4, 0.546, 24}  & \parbox{3cm}{280}& \parbox{3cm}{0.675}\\
\hline
\end{tabular}%
}
\end{adjustbox}
\end{table*}
We also explored whether the use of language affects the perception of the six poles of the three traits by employing the Mann-Whitney U test. This test assessed, for each of the two possible poles of the three personality dimensions, whether there are statistical differences in their perception depending on the experimental condition. Table 3 shows the obtained results. These indicate that the only personality poles significantly affected by the use of language are High Extroversion ($p<0.001$) and Low Agreeableness ($p<0.001$). No significant differences are observed for Low Extroversion, High Agreeableness, or Conscientiousness.
Indeed, the box plot (Figure \ref{fig:boxplot}) and results in Table 2  highlight that the distance between the perception of opposite personality poles (Agreeable-Disagreeable, Extrovert-Introvert) increases in the \textit{Speaking} condition, mainly due to changes in perceived High Extroversion and Low Agreeableness (partially confirming \textit{Hypothesis II}).

However, the fact that there are no statistically significant differences in 4 out of the 6 tested personality poles confirms that the correct selection of parameters and actions is sufficient to display distinct personality traits, even in the case of non-humanoid robots and in the absence of verbal interaction.

\begin{table*}[h]
\centering
\caption{Results of the ANOVA \cite{kim2014analysis} statistical test to investigate if the combination of personality impact on the perception of personality. }
\label{table_example}
\begin{adjustbox}{max width=\textwidth}
\resizebox{16cm}{!}{%
\begin{tabular}{|p{4cm} p{4cm} p{4cm} p{1cm} p{1cm}|}
\hline 
\rowcolor{lightgray}\textbf{Independent Variables}&\parbox{5cm}{ \textbf{Dependent Variable}} & \textbf{Condition} & \textbf{F} & \textbf{p value} \\ 

\parbox{4cm}{\textbf{HE$+$A$+$C}} & \parbox{4cm}{Extroversion}  & \parbox{4cm}{\textit{Not-Speaking}} &  \parbox{1cm}{0.176}  & \parbox{3cm}{0.91}\\

\parbox{4cm}{\textbf{HE$+$A$+$C}} & \parbox{4cm}{Extroversion}  & \parbox{4cm}{\textit{Speaking}} &  \parbox{1cm}{1.48} & \parbox{3cm}{0.270}\\

\parbox{7cm}{\textbf{LE$+$A$+$C}} & \parbox{4cm}{Extroversion}  & \parbox{4cm}{\textit{Not-Speaking}} &  \parbox{1cm}{0.0139}  & \parbox{3cm}{0.998}\\

\parbox{7cm}{\textbf{LE$+$A$+$C}} & \parbox{4cm}{Extroversion}  & \parbox{4cm}{\textit{Speaking}} &  \parbox{1cm}{2.51}  & \parbox{3cm}{0.118}\\

\parbox{7cm}{\textbf{HA$+$E$+$C}} & \parbox{4cm}{Agreeableness}  & \parbox{4cm}{\textit{Not-Speaking}} &  \parbox{1cm}{1.89}  & \parbox{3cm}{0.190}\\

\parbox{7cm}{\textbf{HA$+$E$+$C}} & \parbox{4cm}{Agreeableness}  & \parbox{4cm}{\textit{Speaking}} &  \parbox{1cm}{1.68}  & \parbox{3cm}{0.228}\\

\parbox{7cm}{\textbf{LA$+$E$+$C}} & \parbox{4cm}{Agreeableness}  & \parbox{4cm}{\textit{Not-Speaking}} &  \parbox{1cm}{1.73}  & \parbox{3cm}{0.22}\\

\parbox{7cm}{\textbf{LA$+$E$+$C}} & \parbox{4cm}{Agreeableness}  & \parbox{4cm}{\textit{Speaking}} &  \parbox{1cm}{1.05}  & \parbox{3cm}{0.41}\\

\parbox{7cm}{\textbf{HC$+$E$+$A}} & \parbox{4cm}{Conscientiousness}  & \parbox{4cm}{\textit{Not-Speaking}} &  \parbox{1cm}{0.248}  & \parbox{3cm}{0.861}\\

\parbox{7cm}{\textbf{HC$+$E$+$A}} & \parbox{4cm}{Conscientiousness}  & \parbox{4cm}{\textit{Speaking}} &  \parbox{1cm}{1.01}  & \parbox{3cm}{0.423}\\

\parbox{7cm}{\textbf{LC$+$E$+$A}} & \parbox{4cm}{Conscientiousness}  & \parbox{4cm}{\textit{Not-Speaking}} &  \parbox{1cm}{0.679}  & \parbox{3cm}{0.583}\\

\parbox{7cm}{\textbf{LC$+$E$+$A}} & \parbox{4cm}{Conscientiousness}  & \parbox{4cm}{\textit{Speaking}} &  \parbox{1cm}{1.05}  & \parbox{3cm}{0.41}\\

\hline
\end{tabular}%
}
\end{adjustbox}
\end{table*}
Additionally, the hue of the scatterplot in Figure \ref{fig:boxplot} represents the distribution of each specific pole when combined with the remaining poles of the other two traits. It may be seen how the distribution of points with different colors is spread across the same range of values. Consequently, we decided to investigate if the combination of personality traits had an effect on the perception of specific personality traits. For each condition, for each personality pole we performed an ANOVA analysis \cite{kim2014analysis} (Table 4). We considered as independent variables the four combinations in which a single personality pole can be grouped (e.g. for High Extroversion we considered as independent variables HE$+$HC, HE$+$LC, HE$+$HA, HE$+$LA) and the perceived corresponding personality trait as dependent variable (e.g perceived Extroversion). The results (Table 4) confirm the initial assumption, i.e., combining the CEA traits does not alter the perception of personality.

\begin{table*}[h]
\centering
\caption{Results of the Mann-Whitney U test \cite{nachar2008mann} to investigate if the modulation of one trait is perceived along another OCEAN trait dimention. The considered test Hypothesis is $G1 \neq G2$.}
\label{table_example}
\begin{adjustbox}{max width=\textwidth}
\resizebox{18cm}{!}{%
\begin{tabular}{|p{4cm} p{4cm} p{2cm} p{4cm} p{4cm} p{1cm} p{1cm}|}
\hline 
\rowcolor{lightgray}\textbf{Independent Variables G1, G2}&\parbox{5cm}{ \textbf{Dependent Variable}}& \textbf{Condition} & \textbf{G1: Mean, Mediand, STD, N} & \textbf{G2: Mean, Mediand, STD, N} & \textbf{U } & \textbf{p value} \\ 
%%%%%%%%%%%%%%%% ENS

\parbox{3cm}{\textbf{LE-HE}} & \parbox{3cm}{Agreeableness} & \parbox{3cm}{\textit{Not-Speaking}}  & \parbox{3cm}{3.44, 3.50, 0.851, 24} & \parbox{3cm}{3.38, 3.25,0.680, 24} & \parbox{3cm}{266}& \parbox{3cm}{0.0649}\\  

\parbox{3cm}{\textbf{LE-HE}} & \parbox{3cm}{Conscientiousness} & \parbox{3cm}{\textit{Not-Speaking}}  & \parbox{3cm}{3.58, 3.5, 1.1, 24} & \parbox{3cm}{3.79, 4.0, 0.793, 24 } & \parbox{3cm}{262}& \parbox{3cm}{0.59}\\

\parbox{3cm}{\textbf{LE-HE}} & \parbox{3cm}{Openness} & \parbox{3cm}{\textit{Not-Speaking}}  & \parbox{3cm}{2.71, 3, 0.658, 24} & \parbox{3cm}{2.77, 3, 0.608, 24 } & \parbox{3cm}{286}& \parbox{3cm}{0.974}\\

\parbox{3cm}{\textbf{LE-HE}} & \parbox{3cm}{Emotional Stability} & \parbox{3cm}{\textit{Not-Speaking}}  & \parbox{3cm}{3.85, 4, 0.95, 24} & \parbox{3cm}{3.27, 3.50, 1.123, 24 } & \parbox{3cm}{286}& \parbox{3cm}{0.974}\\

%%%%%%%%%%%%%%%% ES

\parbox{3cm}{\textbf{LE-HE}} & \parbox{3cm}{Agreeableness} & \parbox{3cm}{\textit{Speaking}}  & \parbox{3cm}{3.02, 3.5, 1.026, 24} & \parbox{3cm}{3.08, 3, 1.028, 24} & \parbox{3cm}{304}& \parbox{3cm}{0.867}\\  

\parbox{3cm}{\textbf{LE-HE}} & \parbox{3cm}{Conscientiousness} & \parbox{3cm}{\textit{Speaking}}  & \parbox{3cm}{3.52, 3.5, 0.963, 24} & \parbox{3cm}{3.66, 4, 0.746, 24 } & \parbox{3cm}{305}& \parbox{3cm}{0.882}\\

\parbox{3cm}{\textbf{LE-HE}} & \parbox{3cm}{Openness} & \parbox{3cm}{\textit{Speaking}}  & \parbox{3cm}{2.76, 3, 0.614, 24} & \parbox{3cm}{3.18, 3.5, 0.945, 24 } & \parbox{3cm}{197}& \parbox{3cm}{0.022}\\

\parbox{3cm}{\textbf{LE-HE}} & \parbox{3cm}{Emotional Stability} & \parbox{3cm}{\textit{Speaking}}  & \parbox{3cm}{3.1, 3, 0.99, 24} & \parbox{3cm}{3.34, 3.5, 0.863, 24 } & \parbox{3cm}{268}& \parbox{3cm}{0.386}\\

%%%%%%%%%%%%%%%%%%%%%%%%%% ANS

\parbox{3cm}{\textbf{LA-HA}} & \parbox{3cm}{Extroversion} & \parbox{3cm}{\textit{Not-Speaking}}  & \parbox{3cm}{3, 3, 0.722, 24} & \parbox{3cm}{2.94, 3, 0.742, 24} & \parbox{3cm}{286}& \parbox{3cm}{0.966}\\ 

\parbox{3cm}{\textbf{LA-HA}} & \parbox{3cm}{Conscientiousness} & \parbox{3cm}{\textit{Not-Speaking}}  & \parbox{3cm}{3.75, 4, 0.847, 24} & \parbox{3cm}{3.75, 4, 1.043, 24} & \parbox{3cm}{273}& \parbox{3cm}{0.753}\\ 

\parbox{3cm}{\textbf{LA-HA}} & \parbox{3cm}{Openness} & \parbox{3cm}{\textit{Not-Speaking}}  & \parbox{3cm}{3.02, 3, 0.541, 24} & \parbox{3cm}{2.58, 3, 0.717, 24} & \parbox{3cm}{194}& \parbox{3cm}{0.033}\\ 

\parbox{3cm}{\textbf{LA-HA}} & \parbox{3cm}{Emotional Stability} & \parbox{3cm}{\textit{Not-Speaking}}  & \parbox{3cm}{3.44, 3.75, 1.025, 24} & \parbox{3cm}{3.90, 4.25, 0.989, 24} & \parbox{3cm}{204}& \parbox{3cm}{0.078}\\

%%%%%%%%%%%%%%%%%%%%%%%%%% AS

\parbox{3cm}{\textbf{LA-HA}} & \parbox{3cm}{Extroversion} & \parbox{3cm}{\textit{Speaking}}  & \parbox{3cm}{3.22, 3, 0.542, 24} & \parbox{3cm}{3.58, 4, 0.731, 24} & \parbox{3cm}{187.5}& \parbox{3cm}{0.013}\\ 

\parbox{3cm}{\textbf{LA-HA}} & \parbox{3cm}{Conscientiousness} & \parbox{3cm}{\textit{Speaking}}  & \parbox{3cm}{3.42, 3.5, 1.096, 24} & \parbox{3cm}{3.72, 4, 0.867, 24} & \parbox{3cm}{267}& \parbox{3cm}{0.371}\\ 

\parbox{3cm}{\textbf{LA-HA}} & \parbox{3cm}{Openness} & \parbox{3cm}{\textit{Speaking}}  & \parbox{3cm}{2.98, 3, 0.714, 24} & \parbox{3cm}{2.9, 3, 0.736, 24} & \parbox{3cm}{297.5}& \parbox{3cm}{0.771}\\ 

\parbox{3cm}{\textbf{LA-HA}} & \parbox{3cm}{Emotional Stability} & \parbox{3cm}{\textit{Speaking}}  & \parbox{3cm}{2.28, 2, 0.656, 24} & \parbox{3cm}{3.62, 3.5, 0.794, 24} & \parbox{3cm}{82}& \parbox{3cm}{$<$0.001}\\

%%%%%%%%%%%%%%%%% CNS

\parbox{3cm}{\textbf{LC-HC}} & \parbox{3cm}{Agreeableness} & \parbox{3cm}{\textit{Not-Speaking}}  & \parbox{3cm}{3.08, 3, 0.776, 24} & \parbox{3cm}{3.52, 3.5, 0.744, 24} & \parbox{3cm}{188.5}& \parbox{3cm}{0.038}  \\

\parbox{3cm}{\textbf{LC-HC}} & \parbox{3cm}{Extroversion} & \parbox{3cm}{\textit{Not-Speaking}}  & \parbox{3cm}{2.85, 3, 0.699, 24} & \parbox{3cm}{3.02, 3, 0.878, 24} & \parbox{3cm}{239.5}& \parbox{3cm}{0.311}  \\

\parbox{3cm}{\textbf{LC-HC}} & \parbox{3cm}{Openness} & \parbox{3cm}{\textit{Not-Speaking}}  & \parbox{3cm}{2.69, 3, 0.763, 24} & \parbox{3cm}{2.85, 3, 0.744, 24} & \parbox{3cm}{266}& \parbox{3cm}{0.632}  \\

\parbox{3cm}{\textbf{LC-HC}} & \parbox{3cm}{Emotional Stability} & \parbox{3cm}{\textit{Not-Speaking}}  & \parbox{3cm}{3.08, 3, 1.049, 24} & \parbox{3cm}{3.96, 4, 0.624, 24} & \parbox{3cm}{144}& \parbox{3cm}{0.003}  \\

%%%%%%%%%%%%%%%%% CS

\parbox{3cm}{\textbf{LC-HC}} & \parbox{3cm}{Agreeableness} & \parbox{3cm}{\textit{Speaking}}  & \parbox{3cm}{3.36, 3.50, 0.94, 24} & \parbox{3cm}{2.78, 2.5, 1.109, 24} & \parbox{3cm}{219}& \parbox{3cm}{0.068}  \\

\parbox{3cm}{\textbf{LC-HC}} & \parbox{3cm}{Extroversion} & \parbox{3cm}{\textit{Speaking}}  & \parbox{3cm}{3.48, 3.5, 0.757, 24} & \parbox{3cm}{3.26, 3, 1.091, 24} & \parbox{3cm}{273}& \parbox{3cm}{0.442}  \\

\parbox{3cm}{\textbf{LC-HC}} & \parbox{3cm}{Openness} & \parbox{3cm}{\textit{Speaking}}  & \parbox{3cm}{3, 3, 0.645, 24} & \parbox{3cm}{2.82, 3, 0.69, 24} & \parbox{3cm}{267.5}& \parbox{3cm}{0.372}  \\

\parbox{3cm}{\textbf{LC-HC}} & \parbox{3cm}{Emotional Stability} & \parbox{3cm}{\textit{Speaking}}  & \parbox{3cm}{2.8, 2.5, 0.979, 24} & \parbox{3cm}{3.22, 3.5, 1.011, 24} & \parbox{3cm}{230}& \parbox{3cm}{0.11}  \\
\hline
\end{tabular}%
}
\end{adjustbox}
\end{table*}

We also determined whether a variation in one trait affects the perception of other traits by implementing the Mann-Whitney U test \cite{nachar2008mann} considering as independent variable a variation along one specific traits and as dependent one the perception of the other traits. Table 5 shows the obtained results. 

In the \textit{Speaking} condition, our findings indicate that variations in Extroversion are correlated with perceived variations in Openness ($p = 0.022$). Secondly, changes in Agreeableness are associated with perceptions of Extroversion ($p = 0.013$) and Emotional Stability ($p < 0.001$).  In the \textit{Not-Speaking} condition, Agreeableness correlates with Openness ($p = 0.033$), and Conscientiousness correlates with Emotional Stability ($p = 0.003$)  and Agreeableness ($p=0.038$).

%, which likes to chat, or a silent robot that tends to replace the human without notifying them, are perceived as open to experience. Conversely, a rude robot that expresses itself through language may be also perceived as neurotic. An agreeable robot that communicates verbally is also partially perceived as extroverted (sociable), while a distracted robot, that does not express its careless, is perceived as neurotic.

%We can infer in particular that a sociable and talkative robot, which enjoys chatting, or a robot that replaces humans without notifying them, tends to be perceived as open to experience. Conversely, a rude robot that expresses itself through language might also be seen as neurotic, as well as a distracted robot. An agreeable robot that communicates verbally is somewhat perceived as extroverted.

Summarizing, the analysis of the BFI-10 questionnaire shows how the proposed methodologies may implement perceivable personality traits even in a non-humanoid robot, both through gestures and action choices, and without verbal interaction (\textit{Hypothesis I}). However, as expected, incorporating personality-dependent verbal interaction enables users to perceive these personalities in a (slightly) clearer way  (\textit{Hypothesis II}). %Notably, one of the most interesting findings is that the presence of verbal interaction is not necessary for defining personality traits, even in the case of non-humanoid robots.

The results confirm the architecture's adaptability across different tasks and platforms, highlighting its ability to manage multiple personality traits simultaneously without compromising their perception. The framework demonstrates versatility, supporting implementation on various platforms (including non-humanoid ones) and accommodating diverse tasks and actions, even in the absence of verbal interaction.

\subsection{Perception of Personality - Open questions}
\textit{"Write your impression in interacting with the robot."} We analyzed the responses to this open-ended question to gain a deeper understanding of how the robot's different personality traits were perceived and whether these perceptions led to different feelings experienced by the users. 

To systematically identify the recurring themes in the responses, and its link with the robotic personality, we followed the procedure of thematic analysis outlined by \cite{braun2012thematic}.

We performed the thematic analysis separately for \textit{Not-Speaking} and \textit{Speaking} conditions. The procedure (explained in detail in \cite{braun2012thematic}) resulted in identifying recurring concepts related to \textit{participant feelings during interaction} and \textit{perceptions of the robot} without considering the robot's personality. We divided the obtained recurrent themes by the different poles of each CEA dimension, counting the occurrences of the identified concepts, and comparing opposite poles of the personality traits. Table 6 shows the concepts that appeared at least five times. To minimize the effect of trait combinations, we have eliminated opposite concepts,  counting only the remaining ones (e.g., eliminating "distracted" and "accurate").

In the \textit{Not-Speaking} condition, it is evident that participants accurately perceived the robot's personality through its movement parameters and action choices. This led to positive feelings when interacting with a robot that behaves as expected (e.g., fast, collaborative, precise) and less pleasant feelings towards a robot with low levels of Extroversion, Agreeableness, and Conscientiousness. 

In the \textit{Speaking} condition, where the stimuli are more complex, participants perceived different aspects of the interaction or found it more challenging to describe. This variability made it difficult to draw consistent conclusions. The main difference from the \textit{Not-Speaking} condition is that certain social aspects linked to language emerged (e.g., the robot expressing empathy or being perceived as arrogant), and participants were no longer focused on describing the robot’s movements. This indicates that verbal dimensions may overshadow other behavioral parameters in interpreting personalities (\textit{Hypothesis II}). Finally, a distracted robot, but able to implement verbal interaction, seems to acquire a more social dimension, since participants did not describe the interaction as unpleasant.

\begin{table*}[h]
\centering
\caption{Impact of Robot Personality on Human Feelings and Robot Perception}
\label{tab:multicol}
\begin{adjustbox}{max width=\textwidth}
\resizebox{16cm}{!}{%
\begin{tabular}{cccc}
\hline
\cellcolor{lightgray}\textbf{Trait} & \cellcolor{lightgray}\textbf{Condition} & \cellcolor{lightgray}\textbf{Robot Personality} & \cellcolor{lightgray}\textbf{Human Feelings \& Robot Perception} \\
\hline
\cellcolor{gray!10}\multirow{4}{*}{} & \multirow{2}{*}{\cellcolor{gray!20}\textbf{Not Speaking}} & \cellcolor{gray!20}\textbf{HE} & \cellcolor{gray!20}\begin{tabular}[c]{@{}l@{}}\textbf{Human Feelings:} Pleasant Interaction, Engaging \\ \textbf{Robot Perception:} High Velocity, Visible Movements, Responsive\end{tabular} \\ 
\cellcolor{gray!10}\textbf{Extroversion} & \cellcolor{gray!20}  & \cellcolor{gray!20}\textbf{LE} & \cellcolor{gray!20}\begin{tabular}[c]{@{}l@{}}\textbf{Human Feelings:} Boring, Unpleasant \\ \textbf{Robot Perception:} Slow, Non-Interactive, Minimal Movements\end{tabular} \\ 
\cellcolor{gray!10} & \multirow{2}{*}{\cellcolor{gray!30}\textbf{Speaking}} & \cellcolor{gray!30}\textbf{HE} & \cellcolor{gray!30}\begin{tabular}[c]{@{}l@{}}\textbf{Human Feelings}: Pleasant Interaction, Interesting \\ \textbf{Robot Perception}: - \end{tabular} \\ 
\cellcolor{gray!10} & \cellcolor{gray!30}  & \cellcolor{gray!30} \textbf{LE} & \cellcolor{gray!30}\begin{tabular}[c]{@{}l@{}}\textbf{Human Feelings}: Boring, Unpleasant \\ \textbf{Robot Perception}: Slow, Non-Interactive\end{tabular} \\
\hline

\cellcolor{gray!10}\multirow{4}{*}{} & \multirow{2}{*}{\cellcolor{gray!20}\textbf{Not Speaking}} & \cellcolor{gray!20}\textbf{HA} & \cellcolor{gray!20}\begin{tabular}[c]{@{}l@{}}\textbf{Human Feelings}: Pleasant Interaction \\ \textbf{Robot Perception}: Confident, Collaborative\end{tabular} \\ 
\cellcolor{gray!10}\textbf{Agreeableness} & \cellcolor{gray!20}  & \cellcolor{gray!20}\textbf{LA} & \cellcolor{gray!20}\begin{tabular}[c]{@{}l@{}}\textbf{Human Feelings}: Annoyed, Unpleasant \\ \textbf{Robot Perception}: Non-Interactive, Unpredictable, Replaces Me\end{tabular} \\ 
\cellcolor{gray!10} & \multirow{2}{*}{\cellcolor{gray!30}\textbf{Speaking}} & \cellcolor{gray!30}\textbf{HA} & \cellcolor{gray!30}\begin{tabular}[c]{@{}l@{}}\textbf{Human Feelings}: Pleasant, Empathetic \\ \textbf{Robot Perception}: Gentle, Sociable\end{tabular} \\ 
\cellcolor{gray!10} & \cellcolor{gray!30}  & \cellcolor{gray!30}\textbf{LA} & \cellcolor{gray!30}\begin{tabular}[c]{@{}l@{}}\textbf{Human Feelings}: Judgmental, Engaging \\ \textbf{Robot Perception}: Competitive, Arrogant, Provocative\end{tabular} \\
\hline

\cellcolor{gray!10}\multirow{4}{*}{} & \multirow{2}{*}{\cellcolor{gray!20}\textbf{Not Speaking}} & \cellcolor{gray!20}\textbf{HC} & \cellcolor{gray!20}\begin{tabular}[c]{@{}l@{}}\textbf{Human Feelings}: Pleasant, Competent \\ \textbf{Robot Perception}: Accurate, Interactive\end{tabular} \\ 
\cellcolor{gray!10}\textbf{Conscientiousness} & \cellcolor{gray!20}  & \cellcolor{gray!20}\textbf{LC} & \cellcolor{gray!20}\begin{tabular}[c]{@{}l@{}}\textbf{Human Feelings}: Confused, Unpleasant \\ \textbf{Robot Perception}: Incompetent, Makes Mistakes\end{tabular} \\ 
\cellcolor{gray!10} & \multirow{2}{*}{\cellcolor{gray!30}\textbf{Speaking}} & \cellcolor{gray!30}\textbf{HC} & \cellcolor{gray!30}\begin{tabular}[c]{@{}l@{}}\textbf{Human Feelings}: - \\ \textbf{Robot Perception}: Slow, Competent\end{tabular} \\ 
\cellcolor{gray!10} & \cellcolor{gray!30} & \cellcolor{gray!30}\textbf{LC} & \cellcolor{gray!30}\begin{tabular}[c]{@{}l@{}} \textbf{Human Feelings:} - \\ \textbf{Robot Perception:} Distracted, Lazy, Seeks Help from Humans\end{tabular} \\
\hline
\end{tabular}%
}
\end{adjustbox}
\end{table*}

\section{Conclusions}
\label {conclusions}
In this work, to address the lack of artificial personality implementation in non-humanoid robots, we have integrated a cognitive architecture designed for implementing robotic personality on the Kinova Jaco2 robot. Within this cognitive architecture, personality is defined as a vector in a three-dimensional space comprising Conscientiousness, Agreeableness, and Extroversion. The proposed cognitive architecture allows the robotic personality to influence not only how actions are executed but also the action decision process, memory encoding, prospective thinking, and emotional intelligence.

This paper provides a detailed explanation of the integration of the proposed cognitive architecture into a robotic arm. Our objective was to investigate the effectiveness of the cognitive architecture in implementing robotic personality independently from the platform used and the impact of verbal communication on the perception of personality. To this end, we conducted a between-subjects user study comparing two conditions: \textit{Speaking} and \textit{Not-Speaking}. %The robot participates in a collaborative game with the participants, and its behavior is influenced by its personality.

Results show that the cognitive architecture, regardless of the condition, effectively implements perceivable personality traits in a Kinova Jaco2 arm. Future work will focus on analyzing the social perception of the robot displaying different personality traits. The overarching goal of this research is to adapt the cognitive architecture for real-world applications, aiming to enhance the coexistence of human workers and collaborative robots in the Industry 4.0 sector.

\section{Acknowledgement}
\label{sec:acknowledgement}
This work was carried out within the framework of the project ``RAISE - Robotics and AI for Socio-economic Empowerment`` and has been supported by the European Union - NextGeneration EU, by means of the project PROPER (Curiosity-Driven projects, University of Genoa). 

\section*{Compliance with Ethical Standards}
The authors declare that they have no potential conflicts of interest related to this research.
This research involved human participants interacting with a social robot. No animals were involved in the study.
All participants provided informed consent before engaging in the study, which involved only verbal interaction with the robot. Participants were fully informed about the nature and purpose of the study, and signed consent forms were obtained from each participant.

\addtolength{\textheight}{-0.1cm}
\bibliography{biblio}


\begin{thebibliography}{76}
\ifx \bisbn   \undefined \def \bisbn  #1{ISBN #1}\fi
\ifx \binits  \undefined \def \binits#1{#1}\fi
\ifx \bauthor  \undefined \def \bauthor#1{#1}\fi
\ifx \batitle  \undefined \def \batitle#1{#1}\fi
\ifx \bjtitle  \undefined \def \bjtitle#1{#1}\fi
\ifx \bvolume  \undefined \def \bvolume#1{\textbf{#1}}\fi
\ifx \byear  \undefined \def \byear#1{#1}\fi
\ifx \bissue  \undefined \def \bissue#1{#1}\fi
\ifx \bfpage  \undefined \def \bfpage#1{#1}\fi
\ifx \blpage  \undefined \def \blpage #1{#1}\fi
\ifx \burl  \undefined \def \burl#1{\textsf{#1}}\fi
\ifx \doiurl  \undefined \def \doiurl#1{\url{https://doi.org/#1}}\fi
\ifx \betal  \undefined \def \betal{\textit{et al.}}\fi
\ifx \binstitute  \undefined \def \binstitute#1{#1}\fi
\ifx \binstitutionaled  \undefined \def \binstitutionaled#1{#1}\fi
\ifx \bctitle  \undefined \def \bctitle#1{#1}\fi
\ifx \beditor  \undefined \def \beditor#1{#1}\fi
\ifx \bpublisher  \undefined \def \bpublisher#1{#1}\fi
\ifx \bbtitle  \undefined \def \bbtitle#1{#1}\fi
\ifx \bedition  \undefined \def \bedition#1{#1}\fi
\ifx \bseriesno  \undefined \def \bseriesno#1{#1}\fi
\ifx \blocation  \undefined \def \blocation#1{#1}\fi
\ifx \bsertitle  \undefined \def \bsertitle#1{#1}\fi
\ifx \bsnm \undefined \def \bsnm#1{#1}\fi
\ifx \bsuffix \undefined \def \bsuffix#1{#1}\fi
\ifx \bparticle \undefined \def \bparticle#1{#1}\fi
\ifx \barticle \undefined \def \barticle#1{#1}\fi
\bibcommenthead
\ifx \bconfdate \undefined \def \bconfdate #1{#1}\fi
\ifx \botherref \undefined \def \botherref #1{#1}\fi
\ifx \url \undefined \def \url#1{\textsf{#1}}\fi
\ifx \bchapter \undefined \def \bchapter#1{#1}\fi
\ifx \bbook \undefined \def \bbook#1{#1}\fi
\ifx \bcomment \undefined \def \bcomment#1{#1}\fi
\ifx \oauthor \undefined \def \oauthor#1{#1}\fi
\ifx \citeauthoryear \undefined \def \citeauthoryear#1{#1}\fi
\ifx \endbibitem  \undefined \def \endbibitem {}\fi
\ifx \bconflocation  \undefined \def \bconflocation#1{#1}\fi
\ifx \arxivurl  \undefined \def \arxivurl#1{\textsf{#1}}\fi
\csname PreBibitemsHook\endcsname

%%% 1
\bibitem[\protect\citeauthoryear{Diener and Lucas}{2019}]{diener2019personality}
\begin{botherref}
\oauthor{\bsnm{Diener}, \binits{E.}},
\oauthor{\bsnm{Lucas}, \binits{R.E.}}:
Personality traits.
General psychology: Required reading
\textbf{278}
(2019)
\end{botherref}
\endbibitem

%%% 2
\bibitem[\protect\citeauthoryear{Mount et~al.}{1998}]{mount1998five}
\begin{barticle}
\bauthor{\bsnm{Mount}, \binits{M.K.}},
\bauthor{\bsnm{Barrick}, \binits{M.R.}},
\bauthor{\bsnm{Stewart}, \binits{G.L.}}:
\batitle{Five-factor model of personality and performance in jobs involving interpersonal interactions}.
\bjtitle{Human performance}
\bvolume{11}(\bissue{2-3}),
\bfpage{145}--\blpage{165}
(\byear{1998})
\end{barticle}
\endbibitem

%%% 3
\bibitem[\protect\citeauthoryear{Russell and Wells}{1991}]{russell1991personality}
\begin{barticle}
\bauthor{\bsnm{Russell}, \binits{R.J.}},
\bauthor{\bsnm{Wells}, \binits{P.A.}}:
\batitle{Personality similarity and quality of marriage}.
\bjtitle{Personality and Individual Differences}
\bvolume{12}(\bissue{5}),
\bfpage{407}--\blpage{412}
(\byear{1991})
\end{barticle}
\endbibitem

%%% 4
\bibitem[\protect\citeauthoryear{Correa et~al.}{2010}]{correa2010interacts}
\begin{barticle}
\bauthor{\bsnm{Correa}, \binits{T.}},
\bauthor{\bsnm{Hinsley}, \binits{A.W.}},
\bauthor{\bsnm{De~Zuniga}, \binits{H.G.}}:
\batitle{Who interacts on the web?: The intersection of users’ personality and social media use}.
\bjtitle{Computers in human behavior}
\bvolume{26}(\bissue{2}),
\bfpage{247}--\blpage{253}
(\byear{2010})
\end{barticle}
\endbibitem

%%% 5
\bibitem[\protect\citeauthoryear{Adams}{1995}]{adams1995hitch}
\begin{bbook}
\bauthor{\bsnm{Adams}, \binits{D.}}:
\bbtitle{The Hitch Hiker's Guide to the Galaxy Omnibus}.
\bpublisher{Random House}, \blocation{???}
(\byear{1995})
\end{bbook}
\endbibitem

%%% 6
\bibitem[\protect\citeauthoryear{Peeters et~al.}{2006}]{peeters2006personality}
\begin{barticle}
\bauthor{\bsnm{Peeters}, \binits{M.A.}},
\bauthor{\bsnm{Van~Tuijl}, \binits{H.F.}},
\bauthor{\bsnm{Rutte}, \binits{C.G.}},
\bauthor{\bsnm{Reymen}, \binits{I.M.}}:
\batitle{Personality and team performance: a meta-analysis}.
\bjtitle{European journal of personality}
\bvolume{20}(\bissue{5}),
\bfpage{377}--\blpage{396}
(\byear{2006})
\end{barticle}
\endbibitem

%%% 7
\bibitem[\protect\citeauthoryear{Robert}{2018}]{robert2018personality}
\begin{bchapter}
\bauthor{\bsnm{Robert}, \binits{L.}}:
\bctitle{Personality in the human robot interaction literature: A review and brief critique}.
In: \bbtitle{Robert, LP (2018). Personality in the Human Robot Interaction Literature: A Review and Brief Critique, Proceedings of the 24th Americas Conference on Information Systems, Aug},
pp. \bfpage{16}--\blpage{18}
(\byear{2018})
\end{bchapter}
\endbibitem

%%% 8
\bibitem[\protect\citeauthoryear{Esterwood and Robert}{2021}]{esterwood2021systematic}
\begin{barticle}
\bauthor{\bsnm{Esterwood}, \binits{C.}},
\bauthor{\bsnm{Robert}, \binits{L.P.}}:
\batitle{A systematic review of human and robot personality in health care human-robot interaction}.
\bjtitle{Frontiers in Robotics and AI}
\bvolume{8},
\bfpage{748246}
(\byear{2021})
\end{barticle}
\endbibitem

%%% 9
\bibitem[\protect\citeauthoryear{Robert~Jr et~al.}{2020}]{robert2020review}
\begin{barticle}
\bauthor{\bsnm{Robert~Jr}, \binits{L.P.}},
\bauthor{\bsnm{Alahmad}, \binits{R.}},
\bauthor{\bsnm{Esterwood}, \binits{C.}},
\bauthor{\bsnm{Kim}, \binits{S.}},
\bauthor{\bsnm{You}, \binits{S.}},
\bauthor{\bsnm{Zhang}, \binits{Q.}}, \betal:
\batitle{A review of personality in human--robot interactions}.
\bjtitle{Foundations and Trends{\textregistered} in Information Systems}
\bvolume{4}(\bissue{2}),
\bfpage{107}--\blpage{212}
(\byear{2020})
\end{barticle}
\endbibitem

%%% 10
\bibitem[\protect\citeauthoryear{Rossi et~al.}{2020}]{rossi2020role}
\begin{barticle}
\bauthor{\bsnm{Rossi}, \binits{S.}},
\bauthor{\bsnm{Conti}, \binits{D.}},
\bauthor{\bsnm{Garramone}, \binits{F.}},
\bauthor{\bsnm{Santangelo}, \binits{G.}},
\bauthor{\bsnm{Staffa}, \binits{M.}},
\bauthor{\bsnm{Varrasi}, \binits{S.}},
\bauthor{\bsnm{Di~Nuovo}, \binits{A.}}:
\batitle{The role of personality factors and empathy in the acceptance and performance of a social robot for psychometric evaluations}.
\bjtitle{Robotics}
\bvolume{9}(\bissue{2}),
\bfpage{39}
(\byear{2020})
\end{barticle}
\endbibitem

%%% 11
\bibitem[\protect\citeauthoryear{Paradeda et~al.}{2020}]{paradeda2020persuasion}
\begin{bchapter}
\bauthor{\bsnm{Paradeda}, \binits{R.B.}},
\bauthor{\bsnm{Martinho}, \binits{C.}},
\bauthor{\bsnm{Paiva}, \binits{A.}}:
\bctitle{Persuasion strategies using a social robot in an interactive storytelling scenario}.
In: \bbtitle{Proceedings of the 8th International Conference on Human-Agent Interaction},
pp. \bfpage{69}--\blpage{77}
(\byear{2020})
\end{bchapter}
\endbibitem

%%% 12
\bibitem[\protect\citeauthoryear{D'Angelo et~al.}{2023}]{dangelo23}
\begin{bchapter}
\bauthor{\bsnm{D'Angelo}, \binits{I.}},
\bauthor{\bsnm{Morocutti}, \binits{L.}},
\bauthor{\bsnm{Giunchiglia}, \binits{E.}},
\bauthor{\bsnm{Recchiuto~Carmine}, \binits{T.}},
\bauthor{\bsnm{Sgorbissa}, \binits{A.}}:
\bctitle{Nice and nasty theory of mind for social and antisocial robots}.
In: \bbtitle{ROMAN 2023- The 32nd IEEE International Conference on Robot and Human Interactive Communication}
(\byear{2023}).
\bcomment{IEEE}
\end{bchapter}
\endbibitem

%%% 13
\bibitem[\protect\citeauthoryear{Gockley et~al.}{2005}]{gockley2005designing}
\begin{bchapter}
\bauthor{\bsnm{Gockley}, \binits{R.}},
\bauthor{\bsnm{Bruce}, \binits{A.}},
\bauthor{\bsnm{Forlizzi}, \binits{J.}},
\bauthor{\bsnm{Michalowski}, \binits{M.}},
\bauthor{\bsnm{Mundell}, \binits{A.}},
\bauthor{\bsnm{Rosenthal}, \binits{S.}},
\bauthor{\bsnm{Sellner}, \binits{B.}},
\bauthor{\bsnm{Simmons}, \binits{R.}},
\bauthor{\bsnm{Snipes}, \binits{K.}},
\bauthor{\bsnm{Schultz}, \binits{A.C.}}, \betal:
\bctitle{Designing robots for long-term social interaction}.
In: \bbtitle{2005 IEEE/RSJ International Conference on Intelligent Robots and Systems},
pp. \bfpage{1338}--\blpage{1343}
(\byear{2005}).
\bcomment{IEEE}
\end{bchapter}
\endbibitem

%%% 14
\bibitem[\protect\citeauthoryear{Lim et~al.}{2022}]{Lim2022538}
\begin{bchapter}
\bauthor{\bsnm{Lim}, \binits{M.Y.}},
\bauthor{\bsnm{Lopes}, \binits{J.D.A.}},
\bauthor{\bsnm{Robb}, \binits{D.A.}},
\bauthor{\bsnm{Wilson}, \binits{B.W.}},
\bauthor{\bsnm{Moujahid}, \binits{M.}},
\bauthor{\bsnm{De~Pellegrin}, \binits{E.}},
\bauthor{\bsnm{Hastie}, \binits{H.}}:
\bctitle{We Are All Individuals: The Role of Robot Personality and Human Traits in Trustworthy Interaction}.
(\byear{2022}).
\burl{https://www.scopus.com/inward/record.uri?eid=2-s2.0-85140709926&doi=10.1109%2fRO-MAN53752.2022.9900772&partnerID=40&md5=5e1c4133f42525cbeac01f217bcd7089}
\end{bchapter}
\endbibitem

%%% 15
\bibitem[\protect\citeauthoryear{Aly and Tapus}{2016}]{Aly2016193}
\begin{botherref}
\oauthor{\bsnm{Aly}, \binits{A.}},
\oauthor{\bsnm{Tapus}, \binits{A.}}:
Towards an intelligent system for generating an adapted verbal and nonverbal combined behavior in human–robot interaction.
Autonomous Robots
(2016)
\end{botherref}
\endbibitem

%%% 16
\bibitem[\protect\citeauthoryear{Martins et~al.}{2020}]{martins2020i2e}
\begin{barticle}
\bauthor{\bsnm{Martins}, \binits{P.S.}},
\bauthor{\bsnm{Faria}, \binits{G.}},
\bauthor{\bsnm{Cerqueira}, \binits{J.d.J.F.}}:
\batitle{I2e: A cognitive architecture based on emotions for assistive robotics applications}.
\bjtitle{Electronics}
\bvolume{9}(\bissue{10}),
\bfpage{1590}
(\byear{2020})
\end{barticle}
\endbibitem

%%% 17
\bibitem[\protect\citeauthoryear{Landolfi et~al.}{2023}]{landolfi2023working}
\begin{botherref}
\oauthor{\bsnm{Landolfi}, \binits{L.}},
\oauthor{\bsnm{Pasquali}, \binits{D.}},
\oauthor{\bsnm{Nardelli}, \binits{A.}},
\oauthor{\bsnm{Bernotat}, \binits{J.}},
\oauthor{\bsnm{Rea}, \binits{F.}}:
Working memory based architecture for human-aware navigation in industrial settings
(2023)
\end{botherref}
\endbibitem

%%% 18
\bibitem[\protect\citeauthoryear{Rojas et~al.}{2019}]{rojas2019variational}
\begin{barticle}
\bauthor{\bsnm{Rojas}, \binits{R.A.}},
\bauthor{\bsnm{Garcia}, \binits{M.A.R.}},
\bauthor{\bsnm{Wehrle}, \binits{E.}},
\bauthor{\bsnm{Vidoni}, \binits{R.}}:
\batitle{A variational approach to minimum-jerk trajectories for psychological safety in collaborative assembly stations}.
\bjtitle{IEEE Robotics and Automation Letters}
\bvolume{4}(\bissue{2}),
\bfpage{823}--\blpage{829}
(\byear{2019})
\end{barticle}
\endbibitem

%%% 19
\bibitem[\protect\citeauthoryear{Gualtieri et~al.}{2020}]{gualtieri2020opportunities}
\begin{botherref}
\oauthor{\bsnm{Gualtieri}, \binits{L.}},
\oauthor{\bsnm{Palomba}, \binits{I.}},
\oauthor{\bsnm{Wehrle}, \binits{E.J.}},
\oauthor{\bsnm{Vidoni}, \binits{R.}}:
The opportunities and challenges of sme manufacturing automation: safety and ergonomics in human--robot collaboration.
Industry 4.0 for SMEs: Challenges, opportunities and requirements,
105--144
(2020)
\end{botherref}
\endbibitem

%%% 20
\bibitem[\protect\citeauthoryear{Story et~al.}{2022}]{story2022speed}
\begin{barticle}
\bauthor{\bsnm{Story}, \binits{M.}},
\bauthor{\bsnm{Webb}, \binits{P.}},
\bauthor{\bsnm{Fletcher}, \binits{S.R.}},
\bauthor{\bsnm{Tang}, \binits{G.}},
\bauthor{\bsnm{Jaksic}, \binits{C.}},
\bauthor{\bsnm{Carberry}, \binits{J.}}:
\batitle{Do speed and proximity affect human-robot collaboration with an industrial robot arm?}
\bjtitle{International Journal of Social Robotics}
\bvolume{14}(\bissue{4}),
\bfpage{1087}--\blpage{1102}
(\byear{2022})
\end{barticle}
\endbibitem

%%% 21
\bibitem[\protect\citeauthoryear{Brule et~al.}{2014}]{brule2014robot}
\begin{botherref}
\oauthor{\bsnm{Brule}, \binits{R.}},
\oauthor{\bsnm{Dotsch}, \binits{R.}},
\oauthor{\bsnm{Bijlstra}, \binits{G.}},
\oauthor{\bsnm{Wigboldus}, \binits{D.}},
\oauthor{\bsnm{Haselager}, \binits{P.}}:
Do robot performance and behavioral style affect human trust?
International Journal of Social Robotics
\textbf{6}(4)
(2014)
\end{botherref}
\endbibitem

%%% 22
\bibitem[\protect\citeauthoryear{Porub{\v{c}}inov{\'a} and Fidlerov{\'a}}{2020}]{porubvcinova2020determinants}
\begin{barticle}
\bauthor{\bsnm{Porub{\v{c}}inov{\'a}}, \binits{M.}},
\bauthor{\bsnm{Fidlerov{\'a}}, \binits{H.}}:
\batitle{Determinants of industry 4.0 technology adaption and human-robot collaboration}.
\bjtitle{Research Papers Faculty of Materials Science and Technology Slovak University of Technology}
\bvolume{28}(\bissue{46}),
\bfpage{10}--\blpage{21}
(\byear{2020})
\end{barticle}
\endbibitem

%%% 23
\bibitem[\protect\citeauthoryear{Mou et~al.}{2020}]{mou2020systematic}
\begin{barticle}
\bauthor{\bsnm{Mou}, \binits{Y.}},
\bauthor{\bsnm{Shi}, \binits{C.}},
\bauthor{\bsnm{Shen}, \binits{T.}},
\bauthor{\bsnm{Xu}, \binits{K.}}:
\batitle{A systematic review of the personality of robot: Mapping its conceptualization, operationalization, contextualization and effects}.
\bjtitle{International Journal of Human--Computer Interaction}
\bvolume{36}(\bissue{6}),
\bfpage{591}--\blpage{605}
(\byear{2020})
\end{barticle}
\endbibitem

%%% 24
\bibitem[\protect\citeauthoryear{Nardelli et~al.}{2024a}]{nardelli2024personality}
\begin{bchapter}
\bauthor{\bsnm{Nardelli}, \binits{A.}},
\bauthor{\bsnm{Recchiuto}, \binits{C.T.}},
\bauthor{\bsnm{Sgorbissa}, \binits{A.}}:
\bctitle{Personality- and memory-based software framework for human-robot interaction}.
In: \bbtitle{ICRA 2024- International Conference on Robotics and Automation in PACIFICO Yokohama},
p. \bfpage{6}
(\byear{2024}).
\bcomment{IEEE}
\end{bchapter}
\endbibitem

%%% 25
\bibitem[\protect\citeauthoryear{Nardelli et~al.}{2024b}]{nardelli2024ei}
\begin{bchapter}
\bauthor{\bsnm{Nardelli}, \binits{A.}},
\bauthor{\bsnm{Maccagni}, \binits{G.}},
\bauthor{\bsnm{Minutoli}, \binits{F.}},
\bauthor{\bsnm{Recchiuto}, \binits{C.}},
\bauthor{\bsnm{Sgorbissa}, \binits{A.}}:
\bctitle{Personality- and memory-based framework for emotionally intelligent agents}.
In: \bbtitle{2023 33nd IEEE International Conference on Robot and Human Interactive Communication (RO-MAN)},
p. \bfpage{8}
(\byear{2024}).
\bcomment{IEEE}
\end{bchapter}
\endbibitem

%%% 26
\bibitem[\protect\citeauthoryear{Nardelli et~al.}{2023}]{nardelli2023software}
\begin{bchapter}
\bauthor{\bsnm{Nardelli}, \binits{A.}},
\bauthor{\bsnm{Recchiuto}, \binits{C.}},
\bauthor{\bsnm{Sgorbissa}, \binits{A.}}:
\bctitle{A software framework to encode the psychological dimensions of an artificial agent.}
In: \bbtitle{2023 32nd IEEE International Conference on Robot and Human Interactive Communication (RO-MAN)},
pp. \bfpage{1711}--\blpage{1718}
(\byear{2023}).
\bcomment{IEEE}
\end{bchapter}
\endbibitem

%%% 27
\bibitem[\protect\citeauthoryear{Goldberg}{1981}]{goldberg1981language}
\begin{barticle}
\bauthor{\bsnm{Goldberg}, \binits{L.R.}}:
\batitle{Language and individual differences: The search for universals in personality lexicons}.
\bjtitle{Review of personality and social psychology}
\bvolume{2}(\bissue{1}),
\bfpage{141}--\blpage{165}
(\byear{1981})
\end{barticle}
\endbibitem

%%% 28
\bibitem[\protect\citeauthoryear{Devlin et~al.}{2018}]{devlin2018bert}
\begin{botherref}
\oauthor{\bsnm{Devlin}, \binits{J.}},
\oauthor{\bsnm{Chang}, \binits{M.-W.}},
\oauthor{\bsnm{Lee}, \binits{K.}},
\oauthor{\bsnm{Toutanova}, \binits{K.}}:
Bert: Pre-training of deep bidirectional transformers for language understanding.
arXiv preprint arXiv:1810.04805
(2018)
\end{botherref}
\endbibitem

%%% 29
\bibitem[\protect\citeauthoryear{Bartneck et~al.}{2006}]{bartneck2006use}
\begin{bchapter}
\bauthor{\bsnm{Bartneck}, \binits{C.}},
\bauthor{\bsnm{Reichenbach}, \binits{J.}},
\bauthor{\bsnm{Carpenter}, \binits{J.}}:
\bctitle{Use of praise and punishment in human-robot collaborative teams}.
In: \bbtitle{ROMAN 2006-The 15th IEEE International Symposium on Robot and Human Interactive Communication},
pp. \bfpage{177}--\blpage{182}
(\byear{2006}).
\bcomment{IEEE}
\end{bchapter}
\endbibitem

%%% 30
\bibitem[\protect\citeauthoryear{Riek et~al.}{2009}]{riek2009anthropomorphism}
\begin{bchapter}
\bauthor{\bsnm{Riek}, \binits{L.D.}},
\bauthor{\bsnm{Rabinowitch}, \binits{T.-C.}},
\bauthor{\bsnm{Chakrabarti}, \binits{B.}},
\bauthor{\bsnm{Robinson}, \binits{P.}}:
\bctitle{How anthropomorphism affects empathy toward robots}.
In: \bbtitle{Proceedings of the 4th ACM/IEEE International Conference on Human Robot Interaction},
pp. \bfpage{245}--\blpage{246}
(\byear{2009})
\end{bchapter}
\endbibitem

%%% 31
\bibitem[\protect\citeauthoryear{Mower et~al.}{2007}]{mower2007investigating}
\begin{bchapter}
\bauthor{\bsnm{Mower}, \binits{E.}},
\bauthor{\bsnm{Feil-Seifer}, \binits{D.J.}},
\bauthor{\bsnm{Mataric}, \binits{M.J.}},
\bauthor{\bsnm{Narayanan}, \binits{S.}}:
\bctitle{Investigating implicit cues for user state estimation in human-robot interaction using physiological measurements}.
In: \bbtitle{RO-MAN 2007-The 16th IEEE International Symposium on Robot and Human Interactive Communication},
pp. \bfpage{1125}--\blpage{1130}
(\byear{2007}).
\bcomment{IEEE}
\end{bchapter}
\endbibitem

%%% 32
\bibitem[\protect\citeauthoryear{Tapus et~al.}{2008}]{tapus2008user}
\begin{barticle}
\bauthor{\bsnm{Tapus}, \binits{A.}},
\bauthor{\bsnm{{\c{T}}{\u{a}}pu{\c{s}}}, \binits{C.}},
\bauthor{\bsnm{Matari{\'c}}, \binits{M.J.}}:
\batitle{User—robot personality matching and assistive robot behavior adaptation for post-stroke rehabilitation therapy}.
\bjtitle{Intelligent Service Robotics}
\bvolume{1},
\bfpage{169}--\blpage{183}
(\byear{2008})
\end{barticle}
\endbibitem

%%% 33
\bibitem[\protect\citeauthoryear{Groom et~al.}{2009}]{groom2009my}
\begin{bchapter}
\bauthor{\bsnm{Groom}, \binits{V.}},
\bauthor{\bsnm{Takayama}, \binits{L.}},
\bauthor{\bsnm{Ochi}, \binits{P.}},
\bauthor{\bsnm{Nass}, \binits{C.}}:
\bctitle{I am my robot: The impact of robot-building and robot form on operators}.
In: \bbtitle{Proceedings of the 4th ACM/IEEE International Conference on Human Robot Interaction},
pp. \bfpage{31}--\blpage{36}
(\byear{2009})
\end{bchapter}
\endbibitem

%%% 34
\bibitem[\protect\citeauthoryear{Kim et~al.}{2009}]{kim2009entertainment}
\begin{bchapter}
\bauthor{\bsnm{Kim}, \binits{J.}},
\bauthor{\bsnm{Kwak}, \binits{S.S.}},
\bauthor{\bsnm{Kim}, \binits{M.}}:
\bctitle{Entertainment robot personality design based on basic factors of motions: A case study with rolly}.
In: \bbtitle{RO-MAN 2009-The 18th IEEE International Symposium on Robot and Human Interactive Communication},
pp. \bfpage{803}--\blpage{808}
(\byear{2009}).
\bcomment{IEEE}
\end{bchapter}
\endbibitem

%%% 35
\bibitem[\protect\citeauthoryear{Hendriks et~al.}{2011}]{hendriks2011robot}
\begin{barticle}
\bauthor{\bsnm{Hendriks}, \binits{B.}},
\bauthor{\bsnm{Meerbeek}, \binits{B.}},
\bauthor{\bsnm{Boess}, \binits{S.}},
\bauthor{\bsnm{Pauws}, \binits{S.}},
\bauthor{\bsnm{Sonneveld}, \binits{M.}}:
\batitle{Robot vacuum cleaner personality and behavior}.
\bjtitle{International Journal of Social Robotics}
\bvolume{3},
\bfpage{187}--\blpage{195}
(\byear{2011})
\end{barticle}
\endbibitem

%%% 36
\bibitem[\protect\citeauthoryear{Luo et~al.}{2022}]{Luo2022}
\begin{botherref}
\oauthor{\bsnm{Luo}, \binits{L.}},
\oauthor{\bsnm{Ogawa}, \binits{K.}},
\oauthor{\bsnm{Peebles}, \binits{G.}},
\oauthor{\bsnm{Ishiguro}, \binits{H.}}:
Towards a personality ai for robots: Potential colony capacity of a goal-shaped generative personality model when used for expressing personalities via non-verbal behaviour of humanoid robots.
Frontiers in Robotics and AI
(2022)
\end{botherref}
\endbibitem

%%% 37
\bibitem[\protect\citeauthoryear{Andriella et~al.}{2022}]{andriella2022know}
\begin{barticle}
\bauthor{\bsnm{Andriella}, \binits{A.}},
\bauthor{\bsnm{Huertas-Garcia}, \binits{R.}},
\bauthor{\bsnm{Forgas-Coll}, \binits{S.}},
\bauthor{\bsnm{Torras}, \binits{C.}},
\bauthor{\bsnm{Aleny{\`a}}, \binits{G.}}:
\batitle{“i know how you feel”: The importance of interaction style on users’ acceptance in an entertainment scenario}.
\bjtitle{Interaction Studies}
\bvolume{23}(\bissue{1}),
\bfpage{21}--\blpage{57}
(\byear{2022})
\end{barticle}
\endbibitem

%%% 38
\bibitem[\protect\citeauthoryear{Vernon}{2014}]{vernon2014artificial}
\begin{bbook}
\bauthor{\bsnm{Vernon}, \binits{D.}}:
\bbtitle{Artificial Cognitive Systems: A Primer}.
\bpublisher{MIT Press}, \blocation{???}
(\byear{2014})
\end{bbook}
\endbibitem

%%% 39
\bibitem[\protect\citeauthoryear{Kotseruba and Tsotsos}{2020}]{kotseruba202040}
\begin{barticle}
\bauthor{\bsnm{Kotseruba}, \binits{I.}},
\bauthor{\bsnm{Tsotsos}, \binits{J.K.}}:
\batitle{40 years of cognitive architectures: core cognitive abilities and practical applications}.
\bjtitle{Artificial Intelligence Review}
\bvolume{53}(\bissue{1}),
\bfpage{17}--\blpage{94}
(\byear{2020})
\end{barticle}
\endbibitem

%%% 40
\bibitem[\protect\citeauthoryear{Cabrera-Paniagua and Rubilar-Torrealba}{2022}]{cabrera2022adaptive}
\begin{barticle}
\bauthor{\bsnm{Cabrera-Paniagua}, \binits{D.}},
\bauthor{\bsnm{Rubilar-Torrealba}, \binits{R.}}:
\batitle{Adaptive intelligent autonomous system using artificial somatic markers and big five personality traits}.
\bjtitle{Knowledge-Based Systems}
\bvolume{249},
\bfpage{108995}
(\byear{2022})
\end{barticle}
\endbibitem

%%% 41
\bibitem[\protect\citeauthoryear{Bourgais et~al.}{2020}]{bourgais2020ben}
\begin{botherref}
\oauthor{\bsnm{Bourgais}, \binits{M.}},
\oauthor{\bsnm{Taillandier}, \binits{P.}},
\oauthor{\bsnm{Vercouter}, \binits{L.}}:
Ben: An architecture for the behavior of social agents.
Journal of Artificial Societies and Social Simulation
\textbf{23}(4)
(2020)
\end{botherref}
\endbibitem

%%% 42
\bibitem[\protect\citeauthoryear{}{2024}]{kinova1}
\begin{botherref}
Kinova,"https://assistive.kinovarobotics.com/product/jaco-robotic-arm".
[Online; accessed 2024-08-21]
(2024).
\url{https://assistive.kinovarobotics.com/product/jaco-robotic-arm}
\end{botherref}
\endbibitem

%%% 43
\bibitem[\protect\citeauthoryear{}{2023}]{morphcast}
\begin{botherref}
MorphCast,"https://www.morphcast.com/".
[Online; accessed 2024-06-01]
(2023).
\url{https://www.morphcast.com/}
\end{botherref}
\endbibitem

%%% 44
\bibitem[\protect\citeauthoryear{Ekman}{1992}]{ekman1992there}
\begin{botherref}
\oauthor{\bsnm{Ekman}, \binits{P.}}:
Are there basic emotions?
(1992)
\end{botherref}
\endbibitem

%%% 45
\bibitem[\protect\citeauthoryear{Hoffmann and Nebel}{2001}]{hoffmann2001ff}
\begin{barticle}
\bauthor{\bsnm{Hoffmann}, \binits{J.}},
\bauthor{\bsnm{Nebel}, \binits{B.}}:
\batitle{The ff planning system: Fast plan generation through heuristic search}.
\bjtitle{Journal of Artificial Intelligence Research}
\bvolume{14},
\bfpage{253}--\blpage{302}
(\byear{2001})
\end{barticle}
\endbibitem

%%% 46
\bibitem[\protect\citeauthoryear{Luo et~al.}{2022}]{luo2022identifying}
\begin{barticle}
\bauthor{\bsnm{Luo}, \binits{L.}},
\bauthor{\bsnm{Ogawa}, \binits{K.}},
\bauthor{\bsnm{Ishiguro}, \binits{H.}}:
\batitle{Identifying personality dimensions for engineering robot personalities in significant quantities with small user groups}.
\bjtitle{Robotics}
\bvolume{11}(\bissue{1}),
\bfpage{28}
(\byear{2022})
\end{barticle}
\endbibitem

%%% 47
\bibitem[\protect\citeauthoryear{V{\"o}lkel et~al.}{2022}]{volkel2022user}
\begin{bchapter}
\bauthor{\bsnm{V{\"o}lkel}, \binits{S.T.}},
\bauthor{\bsnm{Schoedel}, \binits{R.}},
\bauthor{\bsnm{Kaya}, \binits{L.}},
\bauthor{\bsnm{Mayer}, \binits{S.}}:
\bctitle{User perceptions of extraversion in chatbots after repeated use}.
In: \bbtitle{Proceedings of the 2022 CHI Conference on Human Factors in Computing Systems},
pp. \bfpage{1}--\blpage{18}
(\byear{2022})
\end{bchapter}
\endbibitem

%%% 48
\bibitem[\protect\citeauthoryear{Polzehl}{2015}]{polzehl2015personality}
\begin{botherref}
\oauthor{\bsnm{Polzehl}, \binits{T.}}:
Personality in speech.
Assessment and automatic classification
(2015)
\end{botherref}
\endbibitem

%%% 49
\bibitem[\protect\citeauthoryear{De~Raad et~al.}{2010}]{de2010only}
\begin{barticle}
\bauthor{\bsnm{De~Raad}, \binits{B.}},
\bauthor{\bsnm{Barelds}, \binits{D.P.}},
\bauthor{\bsnm{Levert}, \binits{E.}},
\bauthor{\bsnm{Ostendorf}, \binits{F.}},
\bauthor{\bsnm{Mla{\v{c}}i{\'c}}, \binits{B.}},
\bauthor{\bsnm{Blas}, \binits{L.D.}},
\bauthor{\bsnm{H{\v{r}}eb{\'\i}{\v{c}}kov{\'a}}, \binits{M.}},
\bauthor{\bsnm{Szirm{\'a}k}, \binits{Z.}},
\bauthor{\bsnm{Szarota}, \binits{P.}},
\bauthor{\bsnm{Perugini}, \binits{M.}}, \betal:
\batitle{Only three factors of personality description are fully replicable across languages: a comparison of 14 trait taxonomies.}
\bjtitle{Journal of personality and social psychology}
\bvolume{98}(\bissue{1}),
\bfpage{160}
(\byear{2010})
\end{barticle}
\endbibitem

%%% 50
\bibitem[\protect\citeauthoryear{Saucier and Ostendorf}{1999}]{saucier1999hierarchical}
\begin{barticle}
\bauthor{\bsnm{Saucier}, \binits{G.}},
\bauthor{\bsnm{Ostendorf}, \binits{F.}}:
\batitle{Hierarchical subcomponents of the big five personality factors: a cross-language replication.}
\bjtitle{Journal of personality and social psychology}
\bvolume{76}(\bissue{4}),
\bfpage{613}
(\byear{1999})
\end{barticle}
\endbibitem

%%% 51
\bibitem[\protect\citeauthoryear{John et~al.}{1988}]{john1988lexical}
\begin{barticle}
\bauthor{\bsnm{John}, \binits{O.P.}},
\bauthor{\bsnm{Angleitner}, \binits{A.}},
\bauthor{\bsnm{Ostendorf}, \binits{F.}}:
\batitle{The lexical approach to personality: A historical review of trait taxonomic research}.
\bjtitle{European journal of Personality}
\bvolume{2}(\bissue{3}),
\bfpage{171}--\blpage{203}
(\byear{1988})
\end{barticle}
\endbibitem

%%% 52
\bibitem[\protect\citeauthoryear{Goldberg}{1990}]{goldberg1990alternative}
\begin{barticle}
\bauthor{\bsnm{Goldberg}, \binits{L.R.}}:
\batitle{An alternative" description of personality": the big-five factor structure.}
\bjtitle{Journal of personality and social psychology}
\bvolume{59}(\bissue{6}),
\bfpage{1216}
(\byear{1990})
\end{barticle}
\endbibitem

%%% 53
\bibitem[\protect\citeauthoryear{Garello et~al.}{2020}]{Garello2020256}
\begin{bchapter}
\bauthor{\bsnm{Garello}, \binits{L.}},
\bauthor{\bsnm{Grella}, \binits{F.}},
\bauthor{\bsnm{Castagnetta}, \binits{S.}},
\bauthor{\bsnm{Bruno}, \binits{B.}},
\bauthor{\bsnm{Recchiuto}, \binits{C.T.}},
\bauthor{\bsnm{Sgorbissa}, \binits{A.}}:
\bctitle{Robot Agreeableness and User Engagement in Verbal Human-Robot Interaction}.
(\byear{2020}).
\burl{https://www.scopus.com/inward/record.uri?eid=2-s2.0-85094314877&doi=10.1109%2fUR49135.2020.9144864&partnerID=40&md5=ce373e80f22e2c9cb0a82e436580a384}
\end{bchapter}
\endbibitem

%%% 54
\bibitem[\protect\citeauthoryear{Craenen et~al.}{2018a}]{craenen2018shaping}
\begin{bchapter}
\bauthor{\bsnm{Craenen}, \binits{B.}},
\bauthor{\bsnm{Deshmukh}, \binits{A.}},
\bauthor{\bsnm{Foster}, \binits{M.E.}},
\bauthor{\bsnm{Vinciarelli}, \binits{A.}}:
\bctitle{Shaping gestures to shape personalities: The relationship between gesture parameters, attributed personality traits and godspeed scores}.
In: \bbtitle{2018 27th IEEE International Symposium on Robot and Human Interactive Communication (RO-MAN)},
pp. \bfpage{699}--\blpage{704}
(\byear{2018}).
\bcomment{IEEE}
\end{bchapter}
\endbibitem

%%% 55
\bibitem[\protect\citeauthoryear{Craenen et~al.}{2018b}]{craenen2018we}
\begin{bchapter}
\bauthor{\bsnm{Craenen}, \binits{B.}},
\bauthor{\bsnm{Deshmukh}, \binits{A.}},
\bauthor{\bsnm{Foster}, \binits{M.E.}},
\bauthor{\bsnm{Vinciarelli}, \binits{A.}}:
\bctitle{Do we really like robots that match our personality? the case of big-five traits, godspeed scores and robotic gestures}.
In: \bbtitle{2018 27th IEEE International Symposium on Robot and Human Interactive Communication (RO-MAN)},
pp. \bfpage{626}--\blpage{631}
(\byear{2018}).
\bcomment{IEEE}
\end{bchapter}
\endbibitem

%%% 56
\bibitem[\protect\citeauthoryear{Delgado-G{\'o}mez et~al.}{2022}]{delgado2022automatic}
\begin{barticle}
\bauthor{\bsnm{Delgado-G{\'o}mez}, \binits{D.}},
\bauthor{\bsnm{Mas{\'o}-Besga}, \binits{A.E.}},
\bauthor{\bsnm{Aguado}, \binits{D.}},
\bauthor{\bsnm{Rubio}, \binits{V.J.}},
\bauthor{\bsnm{Sujar}, \binits{A.}},
\bauthor{\bsnm{Bayona}, \binits{S.}}:
\batitle{Automatic personality assessment through movement analysis}.
\bjtitle{Sensors}
\bvolume{22}(\bissue{10}),
\bfpage{3949}
(\byear{2022})
\end{barticle}
\endbibitem

%%% 57
\bibitem[\protect\citeauthoryear{Gilbert and Wilson}{2007}]{gilbert2007prospection}
\begin{barticle}
\bauthor{\bsnm{Gilbert}, \binits{D.T.}},
\bauthor{\bsnm{Wilson}, \binits{T.D.}}:
\batitle{Prospection: Experiencing the future}.
\bjtitle{Science}
\bvolume{317}(\bissue{5843}),
\bfpage{1351}--\blpage{1354}
(\byear{2007})
\end{barticle}
\endbibitem

%%% 58
\bibitem[\protect\citeauthoryear{DeYoung et~al.}{2010}]{deyoung2010testing}
\begin{barticle}
\bauthor{\bsnm{DeYoung}, \binits{C.G.}},
\bauthor{\bsnm{Hirsh}, \binits{J.B.}},
\bauthor{\bsnm{Shane}, \binits{M.S.}},
\bauthor{\bsnm{Papademetris}, \binits{X.}},
\bauthor{\bsnm{Rajeevan}, \binits{N.}},
\bauthor{\bsnm{Gray}, \binits{J.R.}}:
\batitle{Testing predictions from personality neuroscience: Brain structure and the big five}.
\bjtitle{Psychological science}
\bvolume{21}(\bissue{6}),
\bfpage{820}--\blpage{828}
(\byear{2010})
\end{barticle}
\endbibitem

%%% 59
\bibitem[\protect\citeauthoryear{Roccas et~al.}{2002}]{roccas2002big}
\begin{barticle}
\bauthor{\bsnm{Roccas}, \binits{S.}},
\bauthor{\bsnm{Sagiv}, \binits{L.}},
\bauthor{\bsnm{Schwartz}, \binits{S.H.}},
\bauthor{\bsnm{Knafo}, \binits{A.}}:
\batitle{The big five personality factors and personal values}.
\bjtitle{Personality and social psychology bulletin}
\bvolume{28}(\bissue{6}),
\bfpage{789}--\blpage{801}
(\byear{2002})
\end{barticle}
\endbibitem

%%% 60
\bibitem[\protect\citeauthoryear{Seligman et~al.}{2013}]{seligman2013navigating}
\begin{barticle}
\bauthor{\bsnm{Seligman}, \binits{M.E.}},
\bauthor{\bsnm{Railton}, \binits{P.}},
\bauthor{\bsnm{Baumeister}, \binits{R.F.}},
\bauthor{\bsnm{Sripada}, \binits{C.}}:
\batitle{Navigating into the future or driven by the past}.
\bjtitle{Perspectives on psychological science}
\bvolume{8}(\bissue{2}),
\bfpage{119}--\blpage{141}
(\byear{2013})
\end{barticle}
\endbibitem

%%% 61
\bibitem[\protect\citeauthoryear{Savery et~al.}{2020}]{savery2020emotional}
\begin{botherref}
\oauthor{\bsnm{Savery}, \binits{R.}},
\oauthor{\bsnm{Zahray}, \binits{L.}},
\oauthor{\bsnm{Weinberg}, \binits{G.}}:
Emotional musical prosody for the enhancement of trust in robotic arm communication.
arXiv preprint arXiv:2009.09048
(2020)
\end{botherref}
\endbibitem

%%% 62
\bibitem[\protect\citeauthoryear{La~Viola et~al.}{2022}]{la2022humans}
\begin{bchapter}
\bauthor{\bsnm{La~Viola}, \binits{C.}},
\bauthor{\bsnm{Fiorini}, \binits{L.}},
\bauthor{\bsnm{Mancioppi}, \binits{G.}},
\bauthor{\bsnm{Kim}, \binits{J.}},
\bauthor{\bsnm{Cavallo}, \binits{F.}}:
\bctitle{Humans and robotic arm: Laban movement theory to create emotional connection}.
In: \bbtitle{2022 31st IEEE International Conference on Robot and Human Interactive Communication (RO-MAN)},
pp. \bfpage{566}--\blpage{571}
(\byear{2022}).
\bcomment{IEEE}
\end{bchapter}
\endbibitem

%%% 63
\bibitem[\protect\citeauthoryear{Pease and Lewis}{2015}]{pease2015personality}
\begin{barticle}
\bauthor{\bsnm{Pease}, \binits{C.R.}},
\bauthor{\bsnm{Lewis}, \binits{G.J.}}:
\batitle{Personality links to anger: Evidence for trait interaction and differentiation across expression style}.
\bjtitle{Personality and Individual Differences}
\bvolume{74},
\bfpage{159}--\blpage{164}
(\byear{2015})
\end{barticle}
\endbibitem

%%% 64
\bibitem[\protect\citeauthoryear{DeYoung and Gray.}{2020}]{deyoung2020personality}
\begin{botherref}
\oauthor{\bsnm{DeYoung}, \binits{C.G.}},
\oauthor{\bsnm{Gray.}, \binits{J.R.}}:
Personality neuroscience: Explaining individual differences in affect, behaviour and cognition.,
323--346
(2020)
\end{botherref}
\endbibitem

%%% 65
\bibitem[\protect\citeauthoryear{Hughes et~al.}{2020}]{hughes2020personality}
\begin{barticle}
\bauthor{\bsnm{Hughes}, \binits{D.J.}},
\bauthor{\bsnm{Kratsiotis}, \binits{I.K.}},
\bauthor{\bsnm{Niven}, \binits{K.}},
\bauthor{\bsnm{Holman}, \binits{D.}}:
\batitle{Personality traits and emotion regulation: A targeted review and recommendations.}
\bjtitle{Emotion}
\bvolume{20}(\bissue{1}),
\bfpage{63}
(\byear{2020})
\end{barticle}
\endbibitem

%%% 66
\bibitem[\protect\citeauthoryear{}{2023}]{chatgpt}
\begin{botherref}
Chat-GPT,"https://openai.com/".
[Online; accessed 2024-06-01]
(2023).
\url{https://openai.com/}
\end{botherref}
\endbibitem

%%% 67
\bibitem[\protect\citeauthoryear{}{2023}]{azure}
\begin{botherref}
Azure Speech Service,"https://learn.microsoft.com/en-us/azure/ai-services/speech-service/".
[Online; accessed 2024-06-01]
(2023).
\url{https://learn.microsoft.com/en-us/azure/ai-services/speech-service/}
\end{botherref}
\endbibitem

%%% 68
\bibitem[\protect\citeauthoryear{Robertson et~al.}{1997}]{robertson1997four}
\begin{barticle}
\bauthor{\bsnm{Robertson}, \binits{N.}},
\bauthor{\bsnm{Sanders}, \binits{D.}},
\bauthor{\bsnm{Seymour}, \binits{P.}},
\bauthor{\bsnm{Thomas}, \binits{R.}}:
\batitle{The four-colour theorem}.
\bjtitle{journal of combinatorial theory, Series B}
\bvolume{70}(\bissue{1}),
\bfpage{2}--\blpage{44}
(\byear{1997})
\end{barticle}
\endbibitem

%%% 69
\bibitem[\protect\citeauthoryear{}{2024}]{opencv1}
\begin{botherref}
OpenCV,"https://opencv.org/".
[Online; accessed 2024-08-21]
(2024).
\url{https://opencv.org/}
\end{botherref}
\endbibitem

%%% 70
\bibitem[\protect\citeauthoryear{Collins et~al.}{2009}]{collins2009design}
\begin{barticle}
\bauthor{\bsnm{Collins}, \binits{L.M.}},
\bauthor{\bsnm{Dziak}, \binits{J.J.}},
\bauthor{\bsnm{Li}, \binits{R.}}:
\batitle{Design of experiments with multiple independent variables: a resource management perspective on complete and reduced factorial designs.}
\bjtitle{Psychological methods}
\bvolume{14}(\bissue{3}),
\bfpage{202}
(\byear{2009})
\end{barticle}
\endbibitem

%%% 71
\bibitem[\protect\citeauthoryear{Guido et~al.}{2015}]{guido2015italian}
\begin{barticle}
\bauthor{\bsnm{Guido}, \binits{G.}},
\bauthor{\bsnm{Peluso}, \binits{A.M.}},
\bauthor{\bsnm{Capestro}, \binits{M.}},
\bauthor{\bsnm{Miglietta}, \binits{M.}}:
\batitle{An italian version of the 10-item big five inventory: An application to hedonic and utilitarian shopping values}.
\bjtitle{Personality and Individual Differences}
\bvolume{76},
\bfpage{135}--\blpage{140}
(\byear{2015})
\end{barticle}
\endbibitem

%%% 72
\bibitem[\protect\citeauthoryear{}{2024}]{soscisurvey1}
\begin{botherref}
Sosci Survey,"https://www.soscisurvey.de/de/index".
[Online; accessed 2024-06-01]
(2024).
\url{https://www.soscisurvey.de/de/index}
\end{botherref}
\endbibitem

%%% 73
\bibitem[\protect\citeauthoryear{Cronbach}{1951}]{cronbach1951coefficient}
\begin{barticle}
\bauthor{\bsnm{Cronbach}, \binits{L.J.}}:
\batitle{Coefficient alpha and the internal structure of tests}.
\bjtitle{psychometrika}
\bvolume{16}(\bissue{3}),
\bfpage{297}--\blpage{334}
(\byear{1951})
\end{barticle}
\endbibitem

%%% 74
\bibitem[\protect\citeauthoryear{Nachar et~al.}{2008}]{nachar2008mann}
\begin{barticle}
\bauthor{\bsnm{Nachar}, \binits{N.}}, \betal:
\batitle{The mann-whitney u: A test for assessing whether two independent samples come from the same distribution}.
\bjtitle{Tutorials in quantitative Methods for Psychology}
\bvolume{4}(\bissue{1}),
\bfpage{13}--\blpage{20}
(\byear{2008})
\end{barticle}
\endbibitem

%%% 75
\bibitem[\protect\citeauthoryear{Kim}{2014}]{kim2014analysis}
\begin{barticle}
\bauthor{\bsnm{Kim}, \binits{H.-Y.}}:
\batitle{Analysis of variance (anova) comparing means of more than two groups}.
\bjtitle{Restorative dentistry \& endodontics}
\bvolume{39}(\bissue{1}),
\bfpage{74}--\blpage{77}
(\byear{2014})
\end{barticle}
\endbibitem

%%% 76
\bibitem[\protect\citeauthoryear{Braun and Clarke}{2012}]{braun2012thematic}
\begin{bbook}
\bauthor{\bsnm{Braun}, \binits{V.}},
\bauthor{\bsnm{Clarke}, \binits{V.}}:
\bbtitle{Thematic Analysis.}
\bpublisher{American Psychological Association}, \blocation{???}
(\byear{2012})
\end{bbook}
\endbibitem

\end{thebibliography}
 
\end{document}